\documentclass{article} 
\usepackage{iclr2027_conference,times}
\usepackage{amsmath,amssymb,bm}
\usepackage{graphicx}
\usepackage{booktabs}
\usepackage{xcolor}
\usepackage{hyperref}
\usepackage{url}
\usepackage{placeins}
\usepackage{multirow}
\usepackage{algorithm}
\usepackage{algpseudocode}
\usepackage{colortbl}
\definecolor{tabheader}{RGB}{217, 232, 245}  
\definecolor{tabonline}{RGB}{255, 243, 205}  
\definecolor{tabours}{RGB}{198, 246, 213}    
\hypersetup{
    colorlinks=false,
    pdfborder={0 0 0}
}
\setcitestyle{numbers,square,comma,sort&compress}
\newcommand{\hc}[1]{\raisebox{0.5\baselineskip}[0pt][0pt]{\textbf{#1}}}

\usepackage{amsmath,amsfonts,bm}

\def\eqref#1{equation~\ref{#1}}

\def\1{\bm{1}}

\DeclareMathAlphabet{\mathsfit}{\encodingdefault}{\sfdefault}{m}{sl}
\SetMathAlphabet{\mathsfit}{bold}{\encodingdefault}{\sfdefault}{bx}{n}

\graphicspath{{figures/}}
\title{RoMod: Temporal Routing Modulation via Mixture-of-Experts for Video Anomaly Detection}

\author{%
  \textbf{Chao Huang}\textsuperscript{1} \quad
  \textbf{Pengfei Wei}\textsuperscript{1} \quad
  \textbf{Benfeng Wang}\textsuperscript{1} \quad
  \textbf{Chengliang Liu}\textsuperscript{2} \\[5pt]
  \textbf{Wei Wang}\textsuperscript{1} \quad
  \textbf{Li Shen}\textsuperscript{1} \quad
  \textbf{Wenqi Ren}\textsuperscript{1} \quad
  \textbf{Xiaochun Cao}\textsuperscript{1} \\[8pt]
  \textsuperscript{1}Sun Yat-sen University \qquad
  \textsuperscript{2}University of Macau
}

\iclrfinalcopy 

\begin{document}

\maketitle
\thispagestyle{fancy}
\lhead{Preprint. Under review.}
\begin{abstract}
Intermediate-layer features from multimodal large language models have shown strong potential for video anomaly detection (VAD), yet the origin of their discriminative power remains unclear. We study this question using sparse mixture-of-experts (MoE) models, whose explicit expert structure and sparse activation make their internal computation easier to inspect. With a fully frozen backbone and no additional training, we find that anomaly-related evidence is concentrated in a small set of experts. These experts recur across layers, spontaneously specialize in different anomaly types, and together form a dynamic routing subnetwork. We further show that the output channels most strongly influenced by these experts are also the hidden dimensions that contain the most anomaly-relevant information. Routing statistics can therefore serve as an internal anomaly cue that complements semantic features.
Based on these findings, we propose RoMod, an efficient VAD framework trained with only \(5\%\) of weakly labeled videos. RoMod includes a Routing-Modulated Fusion module, RoMF, and a Routing-aware Temporal Network, RoTN. RoMF uses routing signals to adaptively recalibrate hidden semantic channels. Its design also prevents the routing branch from bypassing semantic features and making predictions on its own. RoTN captures the temporal evolution of anomalies from onset to persistence and termination. Experiments on three benchmarks show that RoMod achieves state-of-the-art performance while running substantially faster than dense backbones of comparable size.
\end{abstract}

\section{Introduction}
\label{sec:intro}
\begin{figure}[t]
\centering
\includegraphics[width=\linewidth]{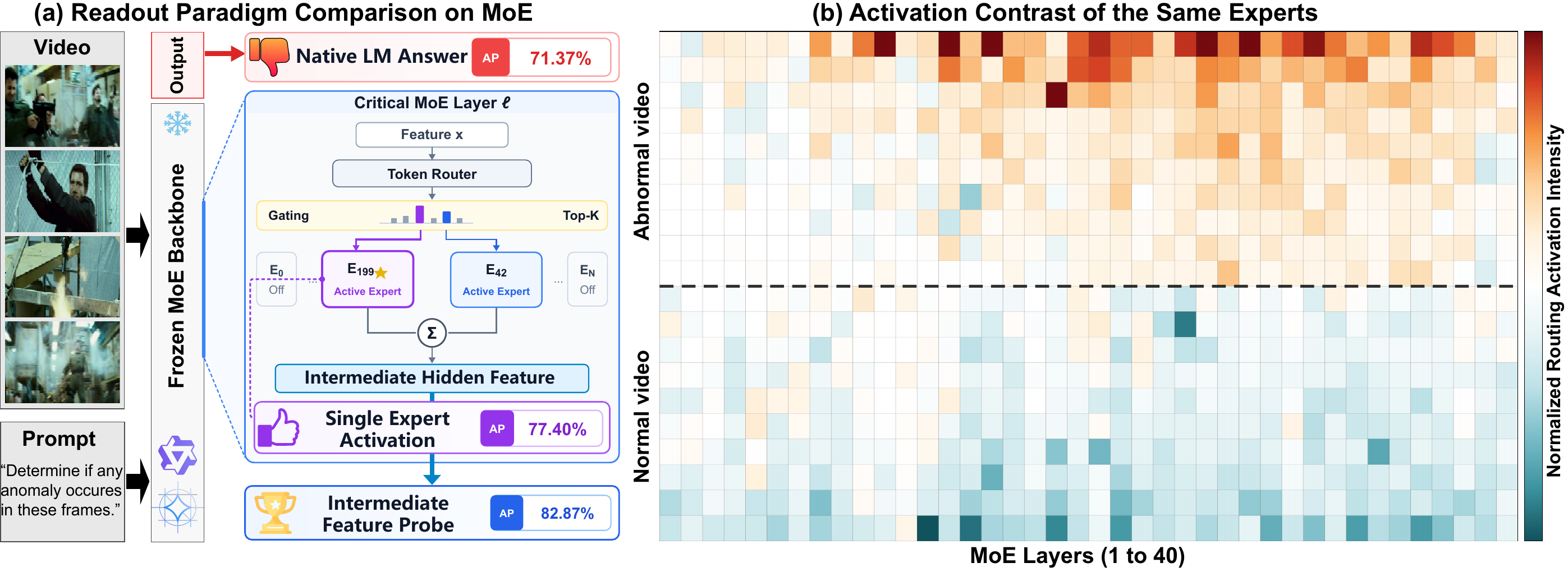}
\caption{Observations on a frozen MoE backbone on XD-Violence test set. 
\textbf{(a)} Comparison of readout APs: native answer ($71.37\%$), hidden probe ($82.87\%$), and a single expert's activation count ($77.40\%$). 
\textbf{(b)} Layer-wise expert activation contrast between normal and abnormal videos (red: high, green: low frequency).}
\label{fig:fig1}
\end{figure}

\paragraph{Latent representation capacity.}
The internal representations of large pre-trained models are often more useful than their generated answers, and across discriminative tasks a lightweight probe on a frozen backbone routinely outperforms the model's own language output~\citep{alain2016probes,belinkov2022probing,burns2022latent,belrose2023tunedlens,cai2025hiprobe}. Dense Transformers, however, activate every parameter for every input, so semantic content, scene layout and capture conditions all end up entangled in the same weights. That makes it hard to say where discriminability emerges, which units carry it, or how it spreads across channels.
\paragraph{Video anomaly detection as a testbed.}
VAD sharpens all three questions. Training labels exist only at the video level while evaluation demands precise temporal localization, so a detector has to separate normal from anomalous clips using its internal representations alone. Violence, accidents and disasters also look nothing alike, and the only thing tying them together is their departure from normal behaviour. In a surveillance stream both accuracy and latency matter, which is where sparse MoE backbones help. Their gating paths split a dense layer into separate, inspectable experts, and only a small subset of those experts fires for any given input~\citep{shazeer2017outrageously,lepikhin2021gshard,fedus2022switch}.
\paragraph{Empirical diagnosis on frozen MoE.}
Figure~\ref{fig:fig1}a compares three ways of reading the same frozen model on XD-Violence. Under the same prompt, the model's textual answer reaches $71.37\%$ AP and a probe of a key hidden layer reaches $82.87\%$ AP. More strikingly, the activation count of a single expert in that layer already reaches $77.40\%$ AP without fitting any parameters or touching the hidden features, which suggests the useful hidden representation is tightly coupled to the model's routing decisions. Across the 40 MoE layers, normal videos stay close to the baseline activation level while anomalous videos drive strong responses in particular middle-to-late experts (Figure~\ref{fig:fig1}b).
\paragraph{The RoMod framework.}
Hidden states describe visual content, whereas routing records which computational pathway the model chose. We combine the two by modulation rather than concatenation, letting the routing vector produce a scale for every hidden channel so that anomaly-related channels are enhanced and background noise damped, without ever replacing the visual-semantic signal. RoMod implements this idea with RoMF and RoTN. Rather than fixing a layer or an expert index in advance, it uses clip-level pseudo-labels from $5\%$ of the weakly labelled videos to select a layer, learn the channel modulation and train the temporal model.
\paragraph{Contributions.}
Our main contributions are summarized as follows.
\begin{itemize}
\item \textbf{A mechanistic analysis.} A forward-only, training-free study of an MoE backbone locates where discriminability arises, identifies the small expert group that carries anomaly semantics, and shows how routing shifts with anomaly type even without category supervision. Expert-channel co-activation statistics tie routing, channel updates and discriminability into one chain.
\item \textbf{Routing-modulated fusion.} RoMF recalibrates hidden channels through a gating-driven multiplicative residual that keeps the semantic and routing branches distinct, while RoTN models the onset, development and end of anomalous states. Together they give a concrete way to exploit internal routing signals from a frozen sparse model.
\item \textbf{A low-resource framework.} A single set of pseudo-labels guides layer selection, modulation and temporal supervision, so neither frame-level labels nor an oracle layer choice is required. Within a $5\%$ data budget, RoMod reaches state-of-the-art results on three benchmarks and supports low-latency streaming detection.
\end{itemize}
\FloatBarrier
\section{Related Work}
\label{sec:related}
\paragraph{Weakly supervised video anomaly detection.}
Frame-level annotation is infeasible for long surveillance footage, so mainstream approaches pair video-level weak labels with multiple-instance learning. A video becomes a bag and a clip an instance, and bag-level supervision is pushed down through ranking losses or top-$k$ aggregation, usually together with temporal smoothing, magnitude constraints or uncertainty modelling to suppress the many normal clips sitting inside anomalous bags~\citep{sultani2018real,tian2021weakly,zhou2023dual,wu2024vadclip}. The representation source has migrated over time, from two-stream action features such as C3D and I3D~\citep{tran2015learning,carreira2017quo}, to vision-language embeddings such as CLIP~\citep{radford2021learning}, and more recently to intermediate-layer hidden states of multimodal large language models~\citep{cai2025hiprobe,zanella2024harnessing,liu2024improved}. That last option brings sizeable gains, yet it is used in a strictly static way. One picks a layer, extracts the hidden state and attaches a temporal head, while the question of which sub-structures the model activated, and with what weights, is discarded entirely.
\paragraph{Expert specialization and internal signals in MoE models.}
Sparse MoE architectures now support a whole generation of large language and multimodal models at activation costs far below dense models of equivalent capacity~\citep{shazeer2017outrageously,lepikhin2021gshard,fedus2022switch}, and their gating mechanism naturally raises the question of whether experts differentiate functionally. Existing analyses concentrate on pure language models, reporting expert preferences for lexical categories, languages or syntactic roles along with regularities in how routing evolves with depth, and their conclusions serve load balancing, expert pruning or routing stability~\citep{zhou2022mixture}. Treating the routing signal itself as a representation for downstream use, particularly for visual temporal discrimination, has to our knowledge not been studied systematically. We neither modify the backbone nor retrain the gate. On a fully frozen model we simply read routing as an internal record of decisions and use it to modulate hidden representations.
\section{Empirical Analysis}
\label{sec:analysis}
All analyses in this section run on a frozen multimodal MoE backbone with no parameter trained. For every clip we record two quantities from the forward pass: the hidden state $h_t^{(l)}$ at layer $l$ and the routing-weight vector $r_t$ produced by the gate, where $t$ indexes clips. Every score reported below therefore comes directly from the pre-trained model rather than from a downstream fitted classifier.
\subsection{Where discriminability lives and which units carry it}
\label{sec:analysis-layer}
\paragraph{Feature evolution exhibits three stages.}
For each layer we measure Fisher separability between normal and anomalous hidden states, comparing the distance between the two class means against the variation inside each class. From L1 to L17 the score sits near zero, because at these depths the representation still describes appearance, texture and local motion, leaving normal and anomalous clips mixed together. It then rises sharply around L18--L27, peaks at L28, and stays high past the plateau. Anomaly semantics thus emerge within a limited depth range instead of accumulating evenly across the network, and once they appear they remain a stable property of the middle-to-late feature space rather than an accident of one layer. Any layer on this plateau is a reasonable candidate, which is what motivates the adaptive selection scheme in Section~\ref{sec:method-readout}. Appendix~\ref{app:layer} gives the setup, the metric derivation and the full curves.
\paragraph{Anomaly semantics concentrate on very few experts.}
\label{sec:analysis-expert}
Separability tells us useful information exists but not which experts carry it. For each layer--expert pair we count how often that expert has a non-zero activation in a clip and use this single number as the anomaly score, with no hidden feature, temporal context or trainable parameter involved. On XD-Violence, expert L26-E199 reaches $91.6\%$ ROC-AUC and $77.40\%$ AP from activation counts alone, far above the random AP baseline of $24.24\%$ and recovering $93\%$ of the AP of the hidden-state probe (Figure~\ref{fig:fig1}a). Transferring the top 20 XD-Violence experts unchanged, the best one still reaches $83.7\%$ ROC-AUC on UCF-Crime and $71.2\%$ AUC on UBnormal, so these experts capture a transferable notion of abnormality rather than a dataset-specific scene bias. Performance declines gradually across the top 20, suggesting the signal is spread over a small expert group instead of residing in one indispensable unit.
\paragraph{Shallow high-response experts reveal a signal misalignment.}
Some high-response experts sit in shallow layers such as L07, L08 and L10, and this does not contradict the poor separability of shallow hidden states. Low-level irregularities can already sway the gate even when the hidden state has not yet expressed that information in a linearly separable form, and it is the feed-forward transforms of later layers that write the discriminative signal into the hidden space. Exactly this mismatch is why routing deserves treatment as a separate feature source.
\subsection{From input-dependent routing to hidden channels}
\label{sec:analysis-routing}
\begin{figure}[t]
\centering
\includegraphics[width=\linewidth]{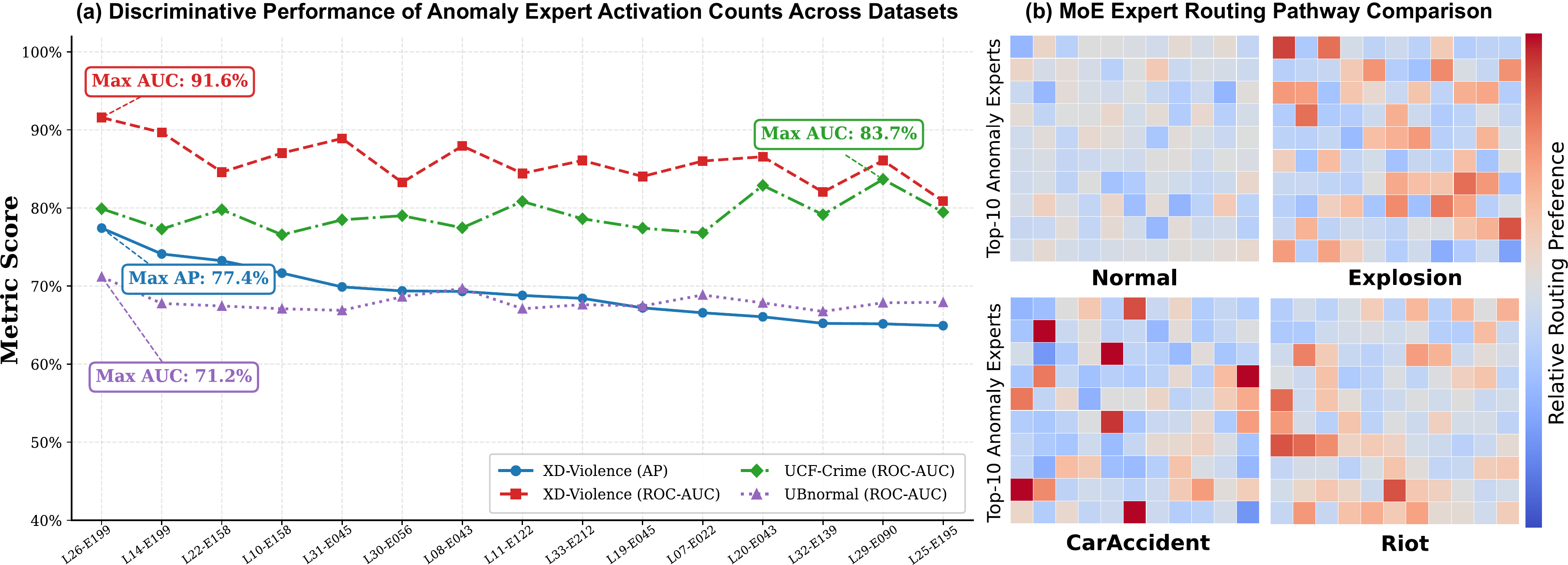}
\caption{Sparse expert specialization and input-dependent routing. \textbf{(a)} Per-expert discriminability from activation counts alone, with no hidden features or trainable parameters. \textbf{(b)} Routing preference over L20--L39 relative to the global average, grouped by event type; cells are comparable horizontally across sub-panels.}
\label{fig:fig2}
\end{figure}

\paragraph{Different anomaly types follow different computational pathways.}
We take the ten most discriminative experts, measure their routing preference relative to the global average across L20--L39, and plot heatmaps grouped by event type (Figure~\ref{fig:fig2}b). Gating parameters are layer-independent, so the same expert index at different depths is a different computational unit and rows are not cross-depth trajectories. What is comparable is the horizontal difference of the same cell across sub-panels, together with the depth-wise clustering of positive offsets. In the normal panel, cells are near zero or slightly negative, meaning that highly discriminative experts merely hold their baseline or get suppressed, whereas in the three anomaly panels the same cells turn strongly positive over large areas. The anomaly types also differ from one another. Explosions concentrate strong preferences into a coherent L25--L32 band, traffic accidents produce sharp isolated responses at L21, L24, L35 and L39, and riots, being crowd-level events, trigger broad gradual enhancement in the shallower L20--L22 range. This differentiation arises without category labels and without fine-tuning, and Appendix~\ref{app:mechanistic} provides multi-seed reproductions along with KL/JS divergence significance tests. Routing is therefore not a static prior bias but an input-dependent signal, so any modulation strategy has to respond dynamically at the clip level.
\paragraph{The causal chain closes at the level of hidden channels.}
\label{sec:analysis-chain}
A standardized channel discriminability analysis locates the hidden dimensions that separate normal and abnormal clips most inside the high plateau. By quantifying the co-activation covariance between routing decisions and residual updates, we find that these discriminative channels align precisely with the hidden dimensions driven most strongly by the high-response anomaly experts, and zeroing those experts in a causal knockout substantially impairs the model's anomaly readout (derivations and knockout results are in Appendix~\ref{app:mechanistic}). The chain therefore closes. An anomalous visual cue drives the gate towards anomaly-specialized experts, whose feed-forward transforms selectively update specific channels and so endow the hidden state with strong linear discriminability. The layer-wise emergence of separability originates in the routing of this sparse specialized group and their coordinated channel writes, and the normal-versus-anomalous contrast in Figure~\ref{fig:fig1}b is its direct manifestation at the gating level.
\subsection{Design constraints implied by the diagnosis}
\label{sec:analysis-design}
\paragraph{Asymmetry between the two signals dictates the fusion operator.}
Hidden states carry the semantic content of scenes, objects and behaviours and form the substrate of discrimination, while routing is an input-dependent selector over computational pathways. Concatenation or summation would erase this asymmetry and let the model bypass semantics to fit the scene directly. We instead let routing generate per-channel scaling factors that strengthen anomaly-bound channels and suppress background interference through a multiplicative residual (Section~\ref{sec:method-romf}). Because expert parameters are independent across layers and effective experts are scattered throughout the network, the modulation module hard-codes no expert index and takes the complete routing vector of the selected layer as input, learning the expert-to-channel mapping from data. The width of the plateau also means that a small amount of weak supervision is enough to locate a layer balancing discriminability against cost, with no oracle layer selection needed (Section~\ref{sec:method-readout}).
\FloatBarrier
\section{Method: RoMod}
\label{sec:method}
\begin{figure}[t]
\centering
\includegraphics[width=\linewidth]{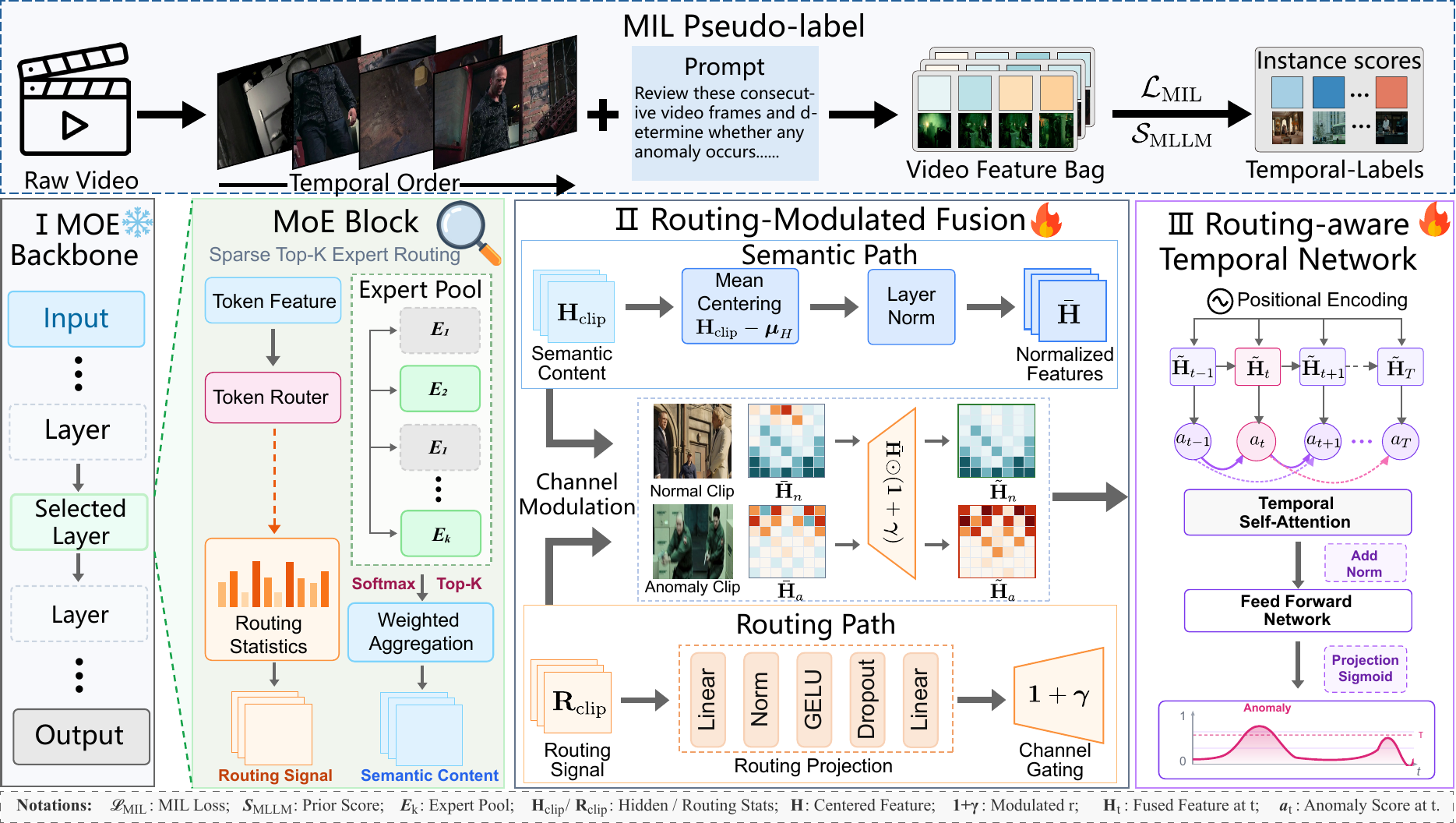}
\vspace{-5pt}
\caption{Overall framework of RoMod. A single forward pass yields the semantic hidden state from the optimal layer $l^{\star}$ and holistic routing statistics across all $L$ layers; RoMF turns the routing vector into per-channel scales that recalibrate the centred hidden state, and RoTN emits clip scores incrementally under causal attention.}
\label{fig:fig3}
\vspace{-3pt}
\end{figure}

Section~\ref{sec:analysis} motivates the three core ingredients of RoMod: an adaptive layer-selection criterion guided by the Fisher plateau, a global routing fingerprint that captures the sparse sub-network, and a gating-modulated fusion operator aligned with how experts write to channels. RoMod (Figure~\ref{fig:fig3}) never touches test ground truth, keeps the multimodal backbone completely frozen, and works in two stages on only $5\%$ of the weakly labelled videos. Stage~I induces clip-level pseudo-labels through lightweight MIL, and those labels then act as the sole supervision in Stage~II, where the channel modulation and the causal temporal network are trained end-to-end.
\subsection{Dual-path signal readout and adaptive layer selection}
\label{sec:method-readout}
\paragraph{Both signals come from a single forward pass.}
We divide each surveillance video into $T$ non-overlapping clips of $T_f=24$ frames, sampling $K=4$ of them uniformly. Each clip goes into an anomaly-query prompt $\mathcal{P}=(\mathcal{P}_{\text{prefix}},\mathcal{P}_{\text{suffix}})$ and then through the frozen MoE backbone, which has $L$ layers with $E$ experts each and routes to the top $K_e$ experts. One forward pass yields two complementary signals for clip $t$. The first is the hidden state $\mathbf{H}_{\text{clip}}^{(l)}(t)$ at candidate layer $l$, a $D$-dimensional vector representing visual semantics. The second is the routing fingerprint $\mathbf{R}_{\text{clip}}(t)$, which concatenates the normalized expert-activation distributions of all $L$ layers. Since the gating computation already produces this information, collecting the fingerprint costs almost nothing. For our 40-layer, 256-expert model it holds $40 \times 256 = 10{,}240$ values and records exactly which sub-networks the input selected.
\paragraph{Stage I: inducing pseudo-labels from a $5\%$ weak-label budget.}
We sample $5\%$ of the weakly labelled training videos to form $\mathcal{D}_{5\%}$. Inside its anomalous
portion, videos are drawn uniformly over the annotated anomaly categories, so rare event types appear in
proportion to the taxonomy rather than to their raw frequency. Each selected video carries only a binary
label $Y$, zero for normal and one for anomalous. A normal video is a pure-negative bag whose every clip
receives a pseudo-label of zero directly, so the induction procedure only has to mine positives inside
the anomalous bags. The frozen language head first provides a zero-shot anomaly score
$\mathcal{S}_{\text{MLLM}}$, which initializes a lightweight category-balanced MIL scorer trained with
the loss $\mathcal{L}_{\text{MIL}}$. Since $l^{\star}$ is itself chosen from the pseudo-labels produced
here, the scorer reads a fixed reference layer $l_{\text{ref}} = L$ instead of $l^{\star}$. We then
suppress false peaks with normal semantic anchors and smooth the result with a Gaussian temporal filter,
which gives a clip-level pseudo-label $\hat{y}_t$ between zero and one. After Stage~I the scorer is
frozen and discarded, and its pseudo-labels become the only supervision used in Stage~II.
Appendix~\ref{app:pseudo} gives the full algorithm and its refinement dynamics.
\paragraph{Adaptive layer selection via between-class pseudo-label separation.}
The Fisher plateau in Section~\ref{sec:analysis-layer} means we need not search every layer exhaustively. We consider the later candidates $\mathcal{L}=\{l_{\min},\dots,L\}$ starting at $l_{\min}=0.5L$, and build a confident anomaly set $\mathcal{A}$ from clips whose pseudo-label is at least $\tau_{\mathrm{hi}}$ together with a normal set $\mathcal{N}$ from clips whose pseudo-label is at most $\tau_{\mathrm{lo}}$. For each layer $l$, the score $G(l)$ below measures the distance between the two mean hidden states after accounting for within-set variation, and we pick the layer that maximizes it with a small preference for shallower depths:
\vspace{-2pt}
\begin{equation}
G(l) = \frac{\big\|\bar{\mathbf{H}}^{(l)}_{\mathcal{A}} - \bar{\mathbf{H}}^{(l)}_{\mathcal{N}}\big\|_2}
{\sqrt{\operatorname{tr}\boldsymbol{\Sigma}^{(l)}_{\mathcal{A}} + \operatorname{tr}\boldsymbol{\Sigma}^{(l)}_{\mathcal{N}}}},
\qquad
l^{\star} = \arg\max_{l \in \mathcal{L}} \Big[ G(l) - \lambda_{\text{reg}}\, \frac{l}{L} \Big].
\label{eq:layer_selection}
\end{equation}
\vspace{-2pt}
The term $\lambda_{\text{reg}}l/L$ is a depth penalty that slightly prefers an earlier layer whenever several layers are similarly separable, which reduces reliance on the final language-oriented layers. Once $l^\star$ is fixed we use its hidden state $\mathbf{h}_t=\mathbf{H}_{\text{clip}}^{(l^\star)}(t)$ as the $D$-dimensional semantic representation of clip $t$, while the routing fingerprint $\mathbf{R}_{\text{clip}}(t)$ always retains information from all $L$ layers.
\subsection{Routing-modulated fusion (RoMF)}
\label{sec:method-romf}
\paragraph{The form of the fusion operator follows the causal chain of Section~\ref{sec:analysis-chain}.}
Since experts influence particular hidden channels, routing should be the thing that adjusts those channels. We first subtract empirical means estimated from $\mathcal{D}_{5\%}$ from both the hidden state and the routing vector, which removes much of the scene and capture-condition bias and yields the centred semantic state $\bar{\mathbf{H}}_t$. A small bottleneck MLP then maps the routing fingerprint to a scale for every hidden channel, with $W_1$ taking the $L\cdot E$ routing values down to a bottleneck of width $d$ and $W_2$ taking it up to the $D$ hidden channels, where $d$ is much smaller than $D$. The resulting channel-wise scale vector $\boldsymbol{\gamma}_t$ then modulates the centred semantic representations as follows:
\begin{equation}
\boldsymbol{\gamma}_t = W_2 \operatorname{Dropout}(\operatorname{GELU}(\operatorname{Norm}(W_1 \mathbf{R}_{\text{clip}}(t)))), \qquad \tilde{\mathbf{H}}_t = (\mathbf{1} + \boldsymbol{\gamma}_t) \odot \bar{\mathbf{H}}_t.
\label{eq:romf_fusion}
\end{equation}
The first expression computes the per-channel scale and the second rescales the centred hidden state element by element, with $\mathbf{1}$ an all-one vector and $\odot$ element-wise multiplication. We initialize $W_2$ to zero, so the module starts out leaving the hidden state untouched and learns the modulation gradually. The modulated state is finally projected to the temporal feature $\mathbf{z}_t$ of dimension $D_{\mathrm{fuse}}$.
\paragraph{The multiplicative form enforces functional decoupling.}
The scale vector $\boldsymbol{\gamma}_t$ can only change the strength of a semantic channel that already exists, never create anomaly evidence on its own. Formally, the derivative of the modulated channel $\tilde{H}_{t,c}$ with respect to its scale $\gamma_{t,c}$ equals the corresponding centred semantic channel $\bar{H}_{t,c}$, so a channel close to zero leaves routing nothing to amplify. Concatenating the semantic vector with a transformed routing vector, or simply adding the two, instead lets the classifier use routing as a shortcut, and because routing signatures are linearly separable that shortcut transfers poorly across domains. Appendix~\ref{app:ablation} verifies this.
\subsection{Causal temporal modelling and Stage-II training (RoTN)}
\label{sec:method-rotn}
\paragraph{The temporal network explicitly models the onset and end of anomalous states.}
The modulated clip features $\{\mathbf{z}_t\}_{t=1}^T$ receive sinusoidal position encodings and pass through a lightweight Transformer encoder with two layers, four heads and a 512-dimensional feature space, where $T$ is the number of clips in the video. A linear layer and a sigmoid turn each output into an anomaly score $a_t$ between zero and one. The attention mask is causal, so when processing clip $t$ the model attends only to clips at times $\tau \le t$, which lets us reuse a rolling KV cache and emit one new score at a time during streaming inference.
\paragraph{Stage II trains only the routing projection branch and the temporal network,}
with the MLLM backbone and the Stage-I pseudo-labels $\hat{y}_t$ kept strictly frozen throughout. The training objective drops any bag-level aggregation term and instead supervises purely at the fine-grained clip level:
\begin{equation}
\mathcal{L} = -\frac{1}{T}\sum_{t=1}^T \Big[ w_{\text{pos}} \hat{y}_t \log(a_t) + (1 - \hat{y}_t)\log(1 - a_t) \Big] 
+ \lambda_{\mathrm{s}} \sum_{t=1}^{T-1} (a_{t+1} - a_t)^2 
+ \lambda_{\mathrm{p}} \frac{1}{T}\sum_{t=1}^T |a_t|,
\label{eq:stage2_loss}
\end{equation}
where $w_{\text{pos}} = N_{\text{neg}} / N_{\text{pos}}$ dynamically counters class imbalance, while the smoothness penalty $\lambda_{\mathrm{s}}$ and the sparsity penalty $\lambda_{\mathrm{p}}$ enforce temporal continuity and event rarity and at the same time damp residual pseudo-label noise from Stage I.
\paragraph{Streaming inference is a single unidirectional pass.}
Clips arrive in temporal order and their semantic features $\mathbf{h}_t$ and network-wide routing fingerprints $\mathbf{R}_{\text{clip}}(t)$ are extracted synchronously, after which RoMF and causal RoTN emit $a_t$ incrementally with no buffering of future clips. A GPU-accelerated 1D Gaussian filter ($\sigma = 8$) then converts clip scores into smooth frame-level probabilities, which serve as the final predictions for evaluation.
\FloatBarrier
\section{Experiments}
\label{sec:exp}
\subsection{Experimental setup}
\label{sec:exp-setup}
We evaluate RoMod on three benchmarks. XD-Violence covers multi-scene violence detection and is reported mainly as frame-level AP, UCF-Crime covers thirteen categories of real-world criminal events and is reported as frame-level ROC-AUC, and UBnormal contains synthetic anomalies with a clear appearance-domain gap from the other two, which makes it a useful probe of cross-domain generalization, also reported as AUC. Training uses only $5\%$ of the weakly labelled videos sampled with category balancing on anomalies (Section~\ref{sec:method-readout}), and all remaining training videos as well as all test ground truth stay unavailable. The default backbone is a frozen Qwen3.6-35B-A3B~\citep{qwen36_35b_a3b} with 40 MoE layers and roughly 2.4B parameters activated per step, and backbone independence is verified on Qwen3VL-30B-A3B~\citep{qwen3vl} and Gemma-4-26B-A4B~\citep{gemma4}. A dense Qwen3.8-27B~\citep{qwen38_27b} serves only as a cost reference in Section~\ref{sec:exp-eff}, since it emits no routing distribution. Trainable parameters amount to the routing projection branch plus a two-layer causal temporal encoder, while clip duration, prompt template, bottleneck width and loss weights are listed in Appendix~\ref{app:impl}. The attention mask of the temporal network decides the deployment mode. Online mode uses causal attention, where each clip sees only history and KV-cached incremental inference is available, whereas offline mode uses bidirectional attention and suits offline retrieval and post-hoc analysis. Main results report both, ablations report offline, and the efficiency analysis reports online.
\subsection{Comparison with existing methods}
\label{sec:exp-main}
\begin{table}[t]
\caption{Comparison on three benchmarks. Data denotes the fraction of training videos used.}
\label{tab:main}
\centering
\footnotesize
\renewcommand{\arraystretch}{1.18}
\setlength{\tabcolsep}{5pt}
\begin{tabular}{ll|c|c|c|cc}
\toprule
\rowcolor{tabheader}
& & & \textbf{UBnormal} & \textbf{UCF-Crime} & \multicolumn{2}{c}{\textbf{XD-Violence}} \\
\rowcolor{tabheader}
\multirow{-2}{*}{\textbf{Method}} & \multirow{-2}{*}{\textbf{Venue}} & \multirow{-2}{*}{\textbf{Data}} & \textbf{AUC (\%)} & \textbf{AUC (\%)} & \textbf{AUC (\%)} & \textbf{AP (\%)} \\
\midrule
UR-DMU~\citep{zhou2023dual}         & AAAI'23    & 100\%  & --    & 86.97 & 94.02 & 81.66 \\
VadCLIP~\citep{wu2024vadclip}        & AAAI'24    & 100\%  & --    & 88.02 & --    & 84.51 \\
VERA~\citep{ye2025vera}              & CVPR'25    & 100\%  & --    & 86.55 & 88.26 & 70.54 \\
$\pi$-VAD~\citep{majhi2025just}      & CVPR'25    & 100\%  & --    & 90.33 & --    & 85.37 \\
RefineVAD~\citep{lee2026refinevad}   & AAAI'26    & 100\%  & --    & 88.92 & --    & 88.66 \\
Anom-$\pi$~\citep{mo2026learning}    & ICML'26    & 100\%  & 78.75 & 84.46 & 93.16 & 72.29 \\
\midrule
LRPO~\citep{leng2026lrpo}            & ICML'26    & 2.5\%  & 75.81 & 85.36 & --    & 73.17 \\
HeadHunt-VAD~\citep{cai2026headhunt} & AAAI'26    & 1\%    & --    & 87.03 & --    & 82.63 \\
HiProbe-VAD~\citep{cai2025hiprobe}   & ACM MM'25  & 1\%    & --    & 86.72 & --    & 82.15 \\
SteerVAD~\citep{cai2026steering}     & ICLR'26    & 1\%    & 72.21 & 87.15 & --    & 83.02 \\
\midrule
EventVAD~\citep{shao2025eventvad}    & ACM MM'25  & 0      & --    & 82.03 & 87.51 & 64.04 \\
Unified-VAD~\citep{lin2026unified}   & NeurIPS'25 & 0      & 69.02 & 84.28 & 91.34 & 68.07 \\
VADTree~\citep{li2026vadtree}        & NeurIPS'25 & 0      & 65.80 & 84.74 & 90.55 & 68.85 \\
PANDA~\citep{yang2026panda}          & NeurIPS'25 & 0      & 75.78 & 84.89 & --    & 70.16 \\
LAVIDA~\citep{dai2026no}             & CVPR'26    & 0      & 76.45 & 82.18 & --    & 90.62 \\
PRISM~\citep{chen2026prism}          & ICML'26    & 0      & --    & 88.85 & 93.42 & 77.48 \\
\midrule
\rowcolor{tabonline}
\textbf{RoMod (online)}  & --   & 5\%    & 77.52 & 88.92 & 95.14 & 90.50 \\
\rowcolor{tabours}
\textbf{RoMod (offline)} & --   & 5\%    & \textbf{79.17} & \textbf{91.32} & \textbf{97.34} & \textbf{92.49} \\
\bottomrule
\end{tabular}
\end{table}
RoMod in offline mode leads on all four reported metrics (Table~\ref{tab:main}). On XD-Violence it reaches $92.49\%$ AP and $97.34\%$ AUC, above LAVIDA's previous best AP of $90.62\%$ and UR-DMU's AUC of $94.02\%$. On UCF-Crime it reaches $91.32\%$ AUC, higher than the $90.33\%$ of the fully supervised $\pi$-VAD, which is notable given that RoMod sees only $5\%$ of the weakly labelled training videos. On UBnormal it obtains $79.17\%$ AUC and again exceeds the full-data Anom-$\pi$ result of $78.75\%$. Since UBnormal has a clear appearance gap from the other two datasets, this outcome supports the analysis in Section~\ref{sec:analysis-expert}, where the routing signal appears to capture abnormal behaviour patterns that transfer beyond a particular scene distribution.
Switching to online inference with causal attention costs about $1.6$, $2.4$ and $2.0$ percentage points on UBnormal, UCF-Crime and XD-Violence respectively, which is expected once the model can no longer inspect future clips. Even so, the online variant stays competitive with most full-data and zero-shot methods, so the accuracy of the approach does not rest on offline temporal context alone.
\subsection{Ablation studies}
\label{sec:exp-ablation}
Table~\ref{tab:ablation} separates the contribution of the two modules. Starting from Hidden-only, RoTN lifts XD-Violence AP from $84.87$ to $86.12$, a modest gain showing that temporal context helps but cannot repair weak clip features. RoMF produces the larger change, raising XD-Violence AP from $84.87$ to $90.95$ and UCF-Crime AUC from $82.90$ to $89.80$, so routing-based channel modulation carries most of the improvement. The two together give the best result, which indicates that feature recalibration and temporal modelling fix different kinds of errors. Corresponding XD-Violence AUC values are $95.28$, $95.80$, $96.85$ and $97.34$.
Table~\ref{tab:budget} examines the supervision budget. Going from $1\%$ to $5\%$ raises XD-Violence AP from $89.20$ to $92.49$, passing through $91.18$ at $2.5\%$. The curve then flattens, since pushing the budget to $10\%$ adds no more than $0.18$ percentage points, probably because extra weakly labelled videos also bring extra pseudo-label noise. We therefore keep $5\%$ as the default, which captures most of the accuracy gain at a far lower annotation cost. Appendix~\ref{app:ablation} contains further fusion and zero-shot-transfer studies.

\begin{table}[t]
\centering
\begin{minipage}[t]{0.535\linewidth}
\caption{Module ablation (offline).}
\label{tab:ablation}
\centering
\footnotesize
\setlength{\tabcolsep}{3pt}
\renewcommand{\arraystretch}{1.18}
\begin{tabular}{ccccc}
\toprule
\rowcolor{tabheader}
\hc{RoMF} & \hc{RoTN} & \shortstack{\textbf{UBnormal}\\\textbf{AUC (\%)}} & \shortstack{\textbf{UCF-Crime}\\\textbf{AUC (\%)}} & \shortstack{\textbf{XD-Violence}\\\textbf{AP (\%)}} \\
\midrule
$\times$     & $\times$     & 72.40 & 82.90 & 84.87 \\
$\times$     & $\checkmark$ & 73.85 & 84.15 & 86.12 \\
$\checkmark$ & $\times$     & 77.92 & 89.80 & 90.95 \\
\rowcolor{tabours}
$\checkmark$ & $\checkmark$ & \textbf{79.17} & \textbf{91.32} & \textbf{92.49} \\
\bottomrule
\end{tabular}
\end{minipage}\hfill
\begin{minipage}[t]{0.44\linewidth}
\caption{Supervision budget (offline).}
\label{tab:budget}
\centering
\footnotesize
\setlength{\tabcolsep}{3.5pt}
\renewcommand{\arraystretch}{1.18}
\begin{tabular}{lccc}
\toprule
\rowcolor{tabheader}
\hc{Data} & \shortstack{\textbf{UBnormal}\\\textbf{AUC (\%)}} & \shortstack{\textbf{UCF-Crime}\\\textbf{AUC (\%)}} & \shortstack{\textbf{XD-Violence}\\\textbf{AP (\%)}} \\
\midrule
1\%          & 75.64 & 88.42 & 89.20 \\
2.5\%        & 77.85 & 90.15 & 91.18 \\
\rowcolor{tabours}
\textbf{5\%} & \textbf{79.17} & \textbf{91.32} & \textbf{92.49} \\
10\%         & 79.35 & 91.46 & 92.61 \\
\bottomrule
\end{tabular}
\end{minipage}
\end{table}
\subsection{Inference efficiency}
\label{sec:exp-eff}
\begin{table}[t]
\caption{Cost and streaming performance in online mode. Activated parameters (Act.) and FLOPs are measured per step; for the dense backbone Act.\ counts all non-embedding parameters.}
\label{tab:eff}
\centering
\footnotesize
\setlength{\tabcolsep}{3.2pt}
\renewcommand{\arraystretch}{1.18}
\resizebox{\linewidth}{!}{
\begin{tabular}{lcccccccc}
\toprule
\rowcolor{tabheader}
\textbf{Backbone} & \textbf{Type} & \textbf{Act. (B)} & \textbf{FLOPs (T)} & \textbf{RoMF (ms)} & \textbf{RoTN (ms)} & \textbf{Thpt. (FPS)} & \parbox[c]{1.8cm}{\centering \textbf{UCF-Crime}\\\textbf{AUC (\%)}} & \parbox[c]{1.9cm}{\centering \textbf{XD-Violence}\\\textbf{AP (\%)}} \\
\midrule
\rowcolor{tabonline}
Qwen3.6-35B-A3B & Sparse & 2.438  & 5.39  & 0.322 & 1.093 & 51.83 & 88.92 & 90.50 \\
Qwen3VL-30B-A3B & Sparse & 2.850  & 6.12  & 0.318 & 1.085 & 36.50 & 87.23 & 88.33 \\
Gemma-4-26B-A4B & Sparse & 3.684  & 10.37 & 0.325 & 0.839 & 39.56 & 86.79 & 87.24 \\
\midrule
Qwen3.8-27B     & Dense  & 24.353 & 54.13 & ---   & ---   & 23.19 & 83.45 & 85.12 \\
\bottomrule
\end{tabular}
}
\end{table}
The 35B-scale backbone (Table~\ref{tab:eff}) activates 2.438B parameters and spends 5.39 TFLOPs per step, reaching 51.83 FPS, well above what real-time surveillance needs. The added modules cost little, with RoMF taking about $0.32$\,ms per step and one incremental RoTN update about $1.09$\,ms on a 10-minute video, since routing values already appear during the forward pass and the causal temporal model reuses its KV cache instead of recomputing earlier clips. The three MoE backbones land in a similar range, suggesting the gain comes from routing structure rather than an implementation detail of one backbone. The dense 27B reference spends $10.0\times$ the FLOPs and falls to 23.19 FPS, a smaller ratio than the FLOPs suggest because the vision encoder and prompt prefill dominate the per-clip cost, and RoMF cannot run on it at all, so its lower scores are a hidden-only reference rather than a method-matched one.
\subsection{Layer-wise gains and qualitative analysis}
\label{sec:exp-layer}
\begin{figure}[t]
\centering
\includegraphics[width=\linewidth]{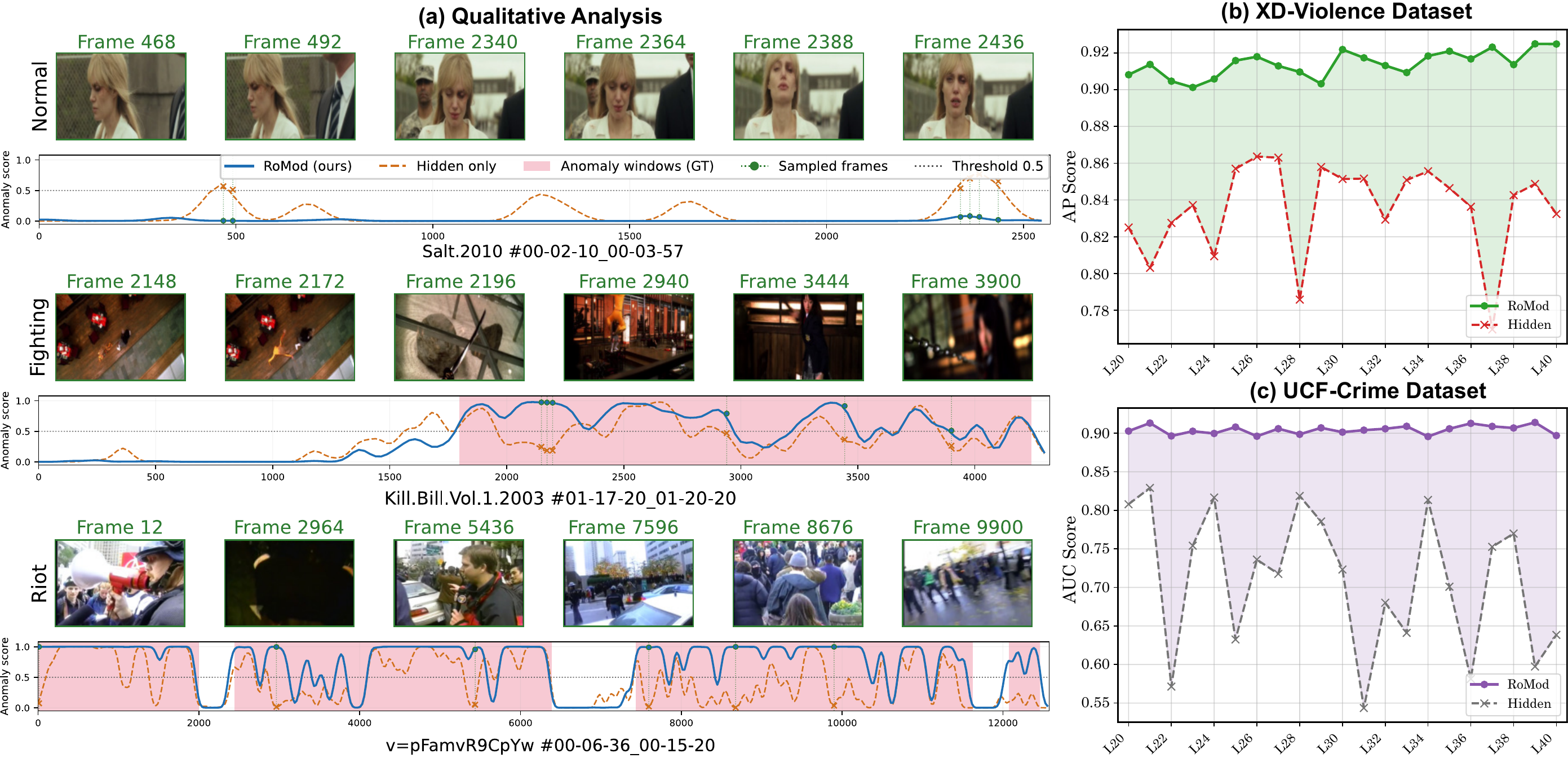}
\caption{Qualitative behaviour and layer-wise gains. \textbf{(a)} Frame-level scores of RoMod (blue) and Hidden-only (orange) on three videos; pink bands are ground-truth windows. \textbf{(b, c)} Layer-wise AP on XD-Violence and AUC on UCF-Crime.}
\label{fig:fig4}
\end{figure}

Figures~\ref{fig:fig4}b and \ref{fig:fig4}c break the plateau of Section~\ref{sec:analysis-layer} down layer by layer from L20 to L40. The Hidden-only baseline swings widely, with UCF-Crime AUC dropping to $0.545$ at L31 and XD-Violence AP troughing at L28 and L38, so a fixed layer choice is unreliable, whereas RoMod improves every tested layer, staying above $0.90$ on XD-Violence and close to $0.90$ on UCF-Crime with its largest gains where Hidden-only is weakest, over $0.35$ at L31. Routing thus compensates for poorly separated hidden states, which makes the pseudo-label-based layer selection of Section~\ref{sec:method-readout} more robust. Figure~\ref{fig:fig4}a shows the same contrast: on the normal video Salt.2010 the baseline throws repeated false alarms above the $0.5$ threshold at shot cuts and illumination changes while RoMod stays near the noise floor, confirming that mean-centring with multiplicative modulation suppresses scene-related channels, and on the fight (Kill.Bill) and riot (Riot) videos RoMod reacts at onset, holds high scores through the window and tracks the multi-segment ground truth, whereas the baseline dips inside anomalous segments and rises inside normal ones. Appendix~\ref{app:vis} presents extended visualizations across diverse accident and crime categories.
\FloatBarrier
\section{Conclusion}
\label{sec:conclusion}
This paper uses a sparse MoE backbone to ask why the intermediate representations of large models
work so well for VAD. Our training-free analysis finds anomaly evidence in a small MoE routing
sub-network whose experts specialize across layers and write to the hidden channels most useful
for discrimination, making routing statistics an internal decision signal that complements visual
semantics. RoMod builds on this with RoMF for routing-driven channel recalibration and RoTN for
causal temporal modelling, reaching state-of-the-art results on three benchmarks from $5\%$ of the
weakly labelled videos while supporting low-latency streaming and suggesting that the internal
dynamics of sparse large models can inform the design of efficient downstream detectors.

\subsection*{AI use statement}
In this work, we used generative AI tools for proofreading and improving text readability. We have not used generative AI tools for scientific idea generation or data simulation. All AI-assisted text was manually reviewed and verified by the authors. The authors take full responsibility for the content of this work.

\subsection*{Ethics statement}
Our study uses only publicly available video anomaly detection benchmarks under their original licences, and no new footage of human subjects was collected. Anomaly detection systems carry surveillance and fairness risks; the frozen-backbone, low-supervision setting studied here does not remove those risks, and deployment should be accompanied by appropriate governance.

\subsection*{Reproducibility statement}
Section~\ref{sec:exp-setup} specifies the benchmarks, backbones and supervision budget. Appendix~\ref{app:mechanistic} and~\ref{app:layer} report the multi-seed reproductions, channel--expert alignment formulations, causal expert intervention, and layer-selection comparisons underlying Section~\ref{sec:analysis}. Appendix~\ref{app:pseudo} details the Stage-I pseudo-label induction procedure and its convergence properties. Complete training recipes, hyper-parameters, and latency breakdowns are provided in Appendix~\ref{app:impl}.

\bibliography{iclr2027_conference}
\bibliographystyle{plainnat}

\clearpage
\appendix

\section{Notation and Conventions}
\label{app:notation}

The appendices reuse several letters across different roles. Table~\ref{tab:notation} fixes the
convention followed throughout; where a symbol in the main text is expanded or disambiguated here,
the correspondence is stated explicitly in the relevant subsection.
\paragraph{Layer indexing convention.}
MoE decoder layers are indexed \emph{one-based} throughout the paper, the appendices and all figure
axes: $l = 1, \dots, L$ with $L = 40$, so the layers are L1 to L40 and ``L$k$'' always denotes the
$k$-th MoE layer. Layer and residual-stream indices coincide under this convention: hidden index $0$ is
the input embedding, and the update emitted by MoE layer $l$ is first observed at hidden index $l$, so
the last layer L40 writes hidden index $40$. Two layer ranges recur and are deliberately distinct. The
\emph{diagnostic band} L20--L39 ($20$ layers) is the window over which the routing-preference heatmaps
(Section~\ref{sec:analysis-routing}), the divergence tests (Appendix~\ref{app:divergence}) and the
knockout intervention (Appendix~\ref{app:knockout}) are computed; its round width keeps the layer
averages in Eq.~\ref{eq:js_kl} and the expert budgets of Appendix~\ref{app:knockout} directly
comparable. The \emph{candidate range} $\mathcal{L} = \{l_{\min}, \dots, L\} = $ L20--L40
($21$ layers, $l_{\min} = 0.5L$) is the layer-selection and layer-wise-evaluation space used in
Section~\ref{sec:method-readout}, Section~\ref{sec:exp-layer} and
Appendix~\ref{app:layer_adaptive}; it additionally contains L40 so that the fixed-last-layer baseline of
Table~\ref{tab:layer_oracle} falls inside it. The three depth stages of Appendix~\ref{app:fisher} are
the shallow-collapse region L1--L17, the emergence band L18--L27, and the high plateau from L28 onward.

\paragraph{Metric conventions.}
Throughout the paper and appendices, \emph{AUC} denotes frame-level ROC-AUC and \emph{AP} denotes
frame-level average precision. Unless stated otherwise, all scores are reported in percent. Where a
quantity is a raw score rather than a percentage (Figures~\ref{fig:fig4}b--c), this is stated in the
caption.

\paragraph{Diagnostic versus deployed quantities.}
Two families of statistics appear in Appendices~\ref{app:mechanistic} and~\ref{app:layer}. Quantities
computed with test-set temporal boundaries ($J(l)$, $S^{(l)}_c$, $\Phi^{(l)}_c$, $z^{(l)}_e$, and the
ground-truth populations $\mathcal{A}, \mathcal{N}$) are \emph{analysis instruments only} and are never
read by RoMod at any stage. Quantities estimated from Stage-I pseudo-labels on $\mathcal{D}_{5\%}$
($G(l)$ and the thresholded sets $\hat{\mathcal{A}}, \hat{\mathcal{N}}$) are the only layer-selection
signals the deployed system uses.

\begin{table}[h]
\centering
\caption{Symbol conventions. Letters that carry more than one role are disambiguated by subscript or
by decoration; the table records which role applies where.}
\label{tab:notation}
\footnotesize
\setlength{\tabcolsep}{5pt}
\renewcommand{\arraystretch}{1.15}
\begin{tabular}{llp{0.50\linewidth}}
\toprule
\textbf{Symbol} & \textbf{Where} & \textbf{Meaning} \\
\midrule
\multicolumn{3}{l}{\textit{Architecture and indices}} \\
$L$, $l$ & everywhere & number of MoE decoder layers; layer index. $L_{\mathrm{moe}}$ is used when
comparing backbones with different depths. \\
$E$, $e$ & everywhere & number of routed experts per layer; expert index. \\
$K_e$ & everywhere & number of experts activated per token (Top-$K_e$ gating), $K_e = 8$. \\
$D$, $c$ & everywhere & hidden width; hidden-channel index. \\
$T$, $t$ & everywhere & number of clips in a video; clip index. \\
$T_f$, $K$ & Appendix~\ref{app:impl} & frames per clip ($24$); frames uniformly sampled per clip ($4$). \\
$\mathcal{L}$ & main text, \ref{app:layer_adaptive} & candidate layer set $\{l_{\min},\dots,L\}$. Losses
are \emph{always} subscripted ($\mathcal{L}_{\mathrm{MIL}}$, $\mathcal{L}_{\mathrm{BCE}}$) and never
written bare. \\
\midrule
\multicolumn{3}{l}{\textit{Populations and sets}} \\
$\mathcal{A}$, $\mathcal{N}$ & \ref{app:alignment}, \ref{app:fisher} & abnormal / normal clip
populations partitioned by \emph{ground-truth} temporal boundaries. \\
$\hat{\mathcal{A}}$, $\hat{\mathcal{N}}$ & \ref{app:layer_adaptive} & confident abnormal / normal sets
obtained by thresholding Stage-I \emph{pseudo-labels}. Written $\mathcal{A},\mathcal{N}$ in
Eq.~\ref{eq:layer_selection}; hats are added here to keep them distinct from the ground-truth
populations. \\
$\mathcal{D}_{5\%}$ & everywhere & the $5\%$ weakly labelled training subset. \\
$\mathcal{D}_y$ & \ref{app:divergence} & clips of anomaly category $y$. \\
$\mathcal{Y}$, $y$ & \ref{app:divergence} & set of annotated anomaly categories; a category. \\
$\mathcal{E}_i$ & \ref{app:multiseed} & Top-$20$ expert set identified in run $i$. \\
$\mathcal{K}^{(l)}$ & \ref{app:knockout} & experts ablated at layer $l$. \\
$\mathcal{Q}_B$ & Alg.~\ref{alg:cmil} & mined positive clip set inside anomalous bag $B$. \\
$\mathcal{I}_{\mathrm{vis}}$, $\mathcal{I}_{\mathrm{text}}$ & \ref{app:prompt_token_agg} & visual and
post-visual suffix token index sets. \\
\midrule
\multicolumn{3}{l}{\textit{Diagnostic statistics (use test ground truth)}} \\
$J(l)$ & \ref{app:fisher} & two-class Fisher separability at layer $l$. \\
$\mathrm{Jac}(\cdot,\cdot)$ & \ref{app:multiseed} & Jaccard similarity between expert sets. Kept
distinct from $J(l)$. \\
$S^{(l)}_c$ & \ref{app:alignment} & channel discriminability score. \\
$z^{(l)}_e$ & \ref{app:alignment}, \ref{app:knockout} & standardized expert anomaly response. \\
$C^{(l)}_{c,e}$ & \ref{app:alignment} & channel--expert co-activation covariance. \\
$\Phi^{(l)}_c$ & \ref{app:alignment} & expert driving contribution on channel $c$. Distinct from the
expert count $E$ and from the expectation $\mathbb{E}[\cdot]$. \\
$K_{\mathrm{ch}}$ & \ref{app:alignment} & size of the Top-$K$ channel sets, $K_{\mathrm{ch}} = 50$. \\
$K_{\mathrm{ko}}$ & \ref{app:knockout} & experts ablated per layer. \\
\midrule
\multicolumn{3}{l}{\textit{Deployed quantities}} \\
$G(l)$ & main text, \ref{app:layer_adaptive} & pseudo-label separability criterion. \\
$\hat{y}_t$ & everywhere & Stage-I clip-level pseudo-label, $\hat{y}_t \in [0,1]$. \\
$\tau_{\mathrm{hi}}$, $\tau_{\mathrm{lo}}$ & \ref{app:layer_adaptive} & pseudo-label confidence
thresholds ($0.7$, $0.1$). \\
$\tau_{\mathrm{dec}}$ & \ref{app:impl}, \ref{app:vis} & decision threshold on $a_t$ ($0.50$). Distinct
from the time index $\tau$ used for causal masking in Section~\ref{sec:method-rotn}. \\
$\lambda_{\mathrm{s}}$, $\lambda_{\mathrm{p}}$, $\lambda_{\mathrm{reg}}$ & everywhere & smoothness,
sparsity, and depth-regularization weights. \\
$\mathcal{G}_\sigma$, $\mathcal{G}^{+}_\sigma$ & everywhere & symmetric and causal half Gaussian
smoothing kernels. \\
\bottomrule
\end{tabular}
\end{table}

\section{Mechanistic Analysis and Causal Attribution Details}
\label{app:mechanistic}

\subsection{Multi-seed Robustness Analysis}
\label{app:multiseed}

To verify that the downstream detection gains are not artefacts of a favourable data split, we conduct
evaluations across $5$ independent runs with distinct random seeds
($\mathrm{seeds} \in \{42, 1024, 2024, 3407, 7777\}$). In each run the $5\%$ weakly supervised training
subset $\mathcal{D}_{5\%}$ is independently re-sampled, and the downstream projection and temporal
modules are re-initialized and trained from scratch.

As shown in Table~\ref{tab:multiseed}, despite the randomness of sampling only $5\%$ of the videos from
complex surveillance distributions, RoMod is stable. The variance induced by re-sampling stays within a
$\sim 1.8$-point envelope below the default split (Seed 42), with standard deviations of $0.49$ to
$0.69$ points. Even on the hardest split (Run 4) RoMod still exceeds the full-data and zero-shot
baselines of Table~\ref{tab:main} by a clear margin.

We further measure Top-$20$ anomaly-expert consistency across runs using the pairwise Jaccard index
\begin{equation}
\mathrm{Jac}(\mathcal{E}_i, \mathcal{E}_j)
= \frac{|\mathcal{E}_i \cap \mathcal{E}_j|}{|\mathcal{E}_i \cup \mathcal{E}_j|},
\label{eq:jaccard}
\end{equation}
where $\mathcal{E}_i$ is the set of Top-$20$ most discriminative experts identified in run $i$. The
average pairwise similarity is $88.2\%$, indicating that the routing sub-network reflects a
sample-invariant property of the backbone rather than noise from a particular split. We write
$\mathrm{Jac}$ rather than $J$ to keep Eq.~\ref{eq:jaccard} distinct from the Fisher separability
$J(l)$ of Appendix~\ref{app:fisher}.

\begin{table}[h]
\centering
\caption{Multi-seed evaluation across $5$ independent runs on three benchmarks using $5\%$ training
data. Seed 42 is the default split reported in the main paper; all other runs fall within a
$\sim 1.8$-point range of it.}
\label{tab:multiseed}
\small
\setlength{\tabcolsep}{6pt}
\begin{tabular}{lcccc}
\toprule
\multirow{2}{*}{\textbf{Seed Index}} & \textbf{UBnormal} & \textbf{UCF-Crime} & \multicolumn{2}{c}{\textbf{XD-Violence}} \\
& \textbf{AUC (\%)} & \textbf{AUC (\%)} & \textbf{AUC (\%)} & \textbf{AP (\%)} \\
\midrule
Run 1 (Seed 42, default) & \textbf{79.17} & \textbf{91.32} & \textbf{97.34} & \textbf{92.49} \\
Run 2 (Seed 1024)        & 78.36 & 90.64 & 96.85 & 91.82 \\
Run 3 (Seed 2024)        & 78.85 & 91.03 & 97.12 & 92.15 \\
Run 4 (Seed 3407)        & 77.42 & 89.65 & 96.08 & 90.73 \\
Run 5 (Seed 7777)        & 78.11 & 90.28 & 96.53 & 91.36 \\
\midrule
\textbf{Min -- Max}      & 77.42 -- 79.17 & 89.65 -- 91.32 & 96.08 -- 97.34 & 90.73 -- 92.49 \\
\textbf{Mean $\pm$ Std}  & \textbf{78.38 $\pm$ 0.68} & \textbf{90.58 $\pm$ 0.66} & \textbf{96.78 $\pm$ 0.49} & \textbf{91.71 $\pm$ 0.69} \\
\bottomrule
\end{tabular}
\end{table}

\subsection{Quantitative Divergence and Significance Hypothesis Testing}
\label{app:divergence}

To test whether the category-dependent routing specialization observed in
Section~\ref{sec:analysis-routing} is statistically significant rather than an artefact of background
noise, we perform non-parametric permutation tests on the expert activation probability simplex across
layers L20--L39. The six categories evaluated here are XD-Violence's annotated event labels; they serve
as the analytical benchmark and are distinct from the six-group prompt whitelist of
Appendix~\ref{app:prompt_token_agg}.

\paragraph{Divergence measures.}
For each anomalous clip $t$ of category $y \in \mathcal{Y}$ we extract its visual routing distributions
across the $20$ middle-to-late layers, $\mathbf{p}_t^{(l)} \in \Delta^{E-1}$, normalized so that
$\sum_{e=1}^{E} p_{t,e}^{(l)} = 1$ with $E = 256$. The category-level empirical mean distribution at
layer $l$ is $\bar{\mathbf{P}}_y^{(l)} = |\mathcal{D}_y|^{-1} \sum_{t \in \mathcal{D}_y}
\mathbf{p}_t^{(l)}$. Inter-category divergence is quantified by the layer-averaged Jensen--Shannon (JS)
and directional Kullback--Leibler (KL) divergences:
\begin{equation}
D_{\mathrm{JS}}(y_i, y_j) = \frac{1}{20} \sum_{l=20}^{39}
\mathrm{JS}\big(\bar{\mathbf{P}}_{y_i}^{(l)} \,\big\|\, \bar{\mathbf{P}}_{y_j}^{(l)}\big),
\qquad
D_{\mathrm{KL}}(y_i \parallel y_j) = \frac{1}{20} \sum_{l=20}^{39}
\mathrm{KL}\big(\bar{\mathbf{P}}_{y_i}^{(l)} \,\big\|\, \bar{\mathbf{P}}_{y_j}^{(l)}\big).
\label{eq:js_kl}
\end{equation}

\begin{figure}[h]
\centering
\includegraphics[width=\linewidth]{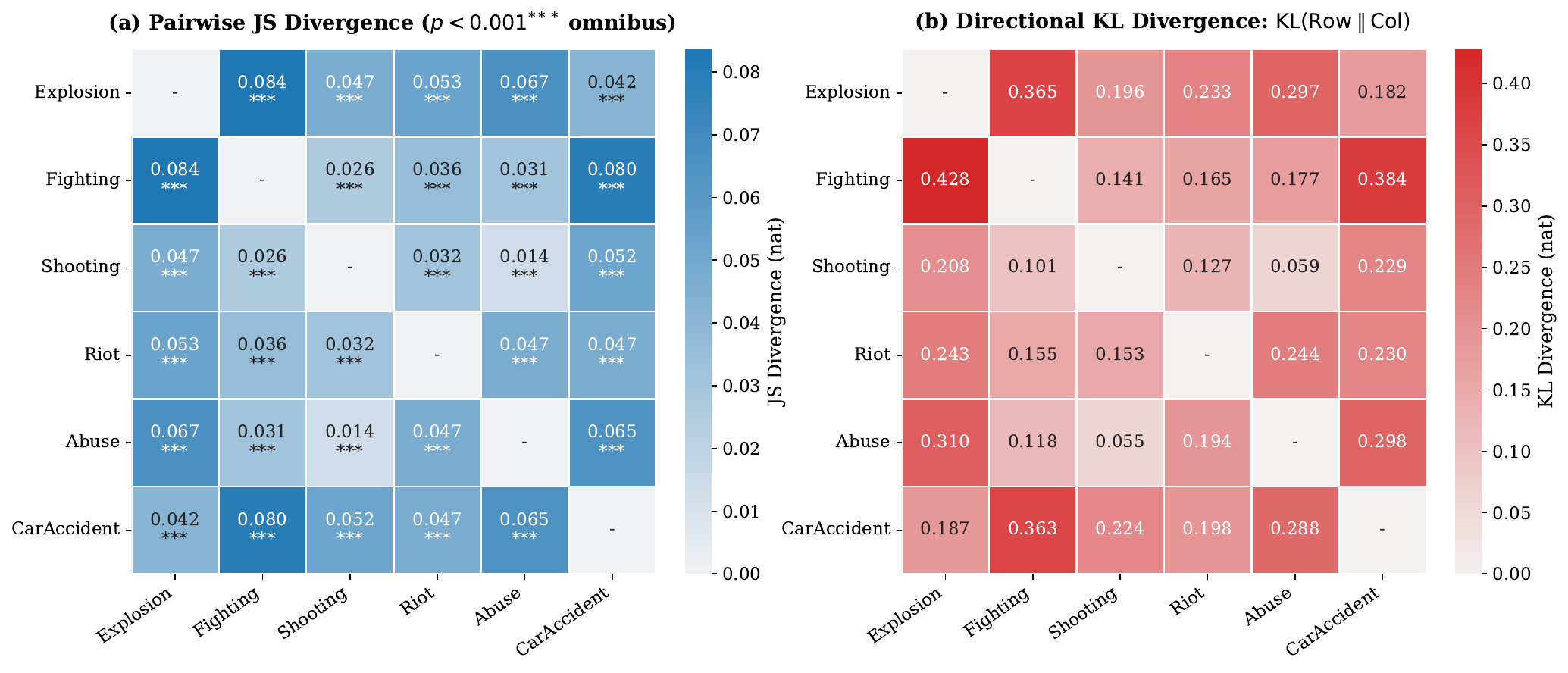}
\caption{\textbf{Statistical hypothesis testing of category-dependent routing specialization on
XD-Violence.}
\textbf{(a)} Layer-averaged pairwise Jensen--Shannon divergence matrix across six anomaly types. Every
pair yields $\hat{p} = 0.0005$ under $2{,}000$ non-parametric permutation tests (marked $^{***}$),
which is the resolution bound of the test.
\textbf{(b)} Directional Kullback--Leibler divergence matrix $\mathrm{KL}(\mathrm{Row} \parallel
\mathrm{Col})$, quantifying the asymmetric information loss when approximating one category's routing
preference by another.}
\label{fig:divergence}
\end{figure}

\paragraph{Permutation protocol and $p$-value convention.}
To test the null hypothesis $H_0$ that MoE routing paths are independent of visual anomaly category, we
run $N = 2{,}000$ Monte Carlo label permutations. All reported $p$-values use the standard
add-one estimator $\hat{p} = (b + 1)/(N + 1)$, where $b$ is the number of permutations whose statistic
equals or exceeds the observed value. With $N = 2{,}000$ this bounds the attainable resolution at
$\hat{p} = 0.0005$, which is the value reported whenever $b = 0$. Two tests are performed:
\begin{enumerate}
\item \textbf{Omnibus test.} Category labels are shuffled across all $23{,}582$ anomalous clips to
build the permutation distribution of the mean off-diagonal JS divergence. The observed global mean is
$\bar{D}_{\mathrm{JS}} = 0.04812$ nat with $b = 0$, hence $\hat{p} = 0.0005$.
\item \textbf{Pairwise tests.} For each pair $(y_i, y_j)$, clips are pooled and randomly partitioned
into equal-sized surrogate groups. As shown in Figure~\ref{fig:divergence}a, every pair attains
$b = 0$ and therefore $\hat{p} = 0.0005$, rejecting $H_0$ at the resolution limit of the test.
\end{enumerate}
We report $\hat{p} = 0.0005$ uniformly rather than mixing it with the looser statement $p < 0.001$;
the two describe the same outcome, and the former states the attainable bound explicitly.

\paragraph{Interpretation.}
Visually distinct anomaly types invoke measurably different sparse computational sub-networks in the
frozen backbone. The largest separation is between the global texture eruptions of \emph{Explosion} and
the localized multi-agent kinematics of \emph{Fighting} ($D_{\mathrm{JS}} = 0.084$ nat). We note that
the tests establish that routing distributions differ across categories; they do not establish that the
activated expert sets are disjoint, and Figure~\ref{fig:divergence}b shows substantial directional
overlap for several pairs.

\subsection{Channel--Expert Correspondence: Formulation and Layer-wise Verification}
\label{app:alignment}

Section~\ref{sec:analysis-chain} claims that the hidden dimensions carrying anomaly discriminability
are coupled to the dimensions written by the anomaly-preferring experts. This subsection gives the
formal statistical formulation, the layer-by-layer validation, and a cross-architecture replication of
that correspondence. All statistics are read at the terminal text token of the anomaly-querying prompt,
matching the readout representation used by the probing experiment in Figure~\ref{fig:fig1}a, and no
trainable parameter is fitted anywhere in this evaluation.

\paragraph{Mathematical formulation.}
the corresponding hidden state $\mathbf{h}^{(l)}_t \in \mathbb{R}^{D}$. Layer and hidden indices
coincide under the one-based convention of Appendix~\ref{app:notation}: the input embedding occupies
hidden index $0$, so the residual update emitted by MoE layer $l$ is first observed at hidden index
$l$, and the last layer L40 writes hidden index $40$.

To isolate dimensions that exhibit genuine distributional separation rather than isolated scaling
artefacts, we define the \emph{channel discriminability score} $S^{(l)}_{c}$ as the product of an
absolute log-variance dispersion ratio and a standardized mean shift (Cohen's $d$):
\begin{equation}
S^{(l)}_{c} =
\underbrace{\left| \log \frac{\sigma^{2}_{\mathcal{A},c}}{\sigma^{2}_{\mathcal{N},c}} \right|}_{\text{dispersion shift}}
\cdot
\underbrace{\frac{\big| \mu_{\mathcal{A},c} - \mu_{\mathcal{N},c} \big|}
{\sqrt{\tfrac{1}{2}\big(\sigma^{2}_{\mathcal{A},c} + \sigma^{2}_{\mathcal{N},c}\big)}}}_{\text{standardized mean shift}},
\qquad c = 1, \dots, D,
\label{eq:chan_score}
\end{equation}
where $\mu_{\cdot,c}$ and $\sigma^{2}_{\cdot,c}$ are the class-conditional empirical mean and variance
of channel $c$ at hidden index $l+1$. Symmetrically, the \emph{expert anomaly response} $z^{(l)}_{e}$
standardizes the routing shift against the global activation variance over all clips:
\begin{equation}
z^{(l)}_{e} = \frac{\bar{r}^{(l)}_{\mathcal{A},e} - \bar{r}^{(l)}_{\mathcal{N},e}}
{\sqrt{\operatorname{Var}_{t \in \mathcal{A} \cup \mathcal{N}}\big[r^{(l)}_{t,e}\big]}},
\qquad e = 1, \dots, E.
\label{eq:expert_resp}
\end{equation}
Conditioned on the anomalous population $\mathcal{A}$, the empirical co-activation covariance between
channel $c$ and expert $e$ is
\begin{equation}
C^{(l)}_{c,e} = \frac{1}{|\mathcal{A}| - 1} \sum_{t \in \mathcal{A}}
\big(h^{(l+1)}_{t,c} - \mu_{\mathcal{A},c}\big)\big(r^{(l)}_{t,e} - \bar{r}^{(l)}_{\mathcal{A},e}\big).
\label{eq:coact}
\end{equation}
Aggregating over anomaly-preferring experts via positive rectification gives the \emph{expert driving
contribution} $\Phi^{(l)}_{c}$:
\begin{equation}
\Phi^{(l)}_{c} = \sum_{e=1}^{E} \big| C^{(l)}_{c,e} \big| \cdot \big[ z^{(l)}_{e} \big]_{+},
\label{eq:driving}
\end{equation}
where $|C^{(l)}_{c,e}|$ accounts for feed-forward writes of either sign and
$[x]_{+} = \max(0, x)$ removes background-suppressed experts. We write $\Phi$ rather than $E$ to keep
Eq.~\ref{eq:driving} distinct from the expert count $E$ and from the expectation
$\mathbb{E}[\cdot]$ used below.

\paragraph{Statistical hypothesis testing.}
Treating the $D$ channels as samples, we test the coupling between $\{\Phi^{(l)}_{c}\}_{c=1}^D$ and
$\{S^{(l)}_{c}\}_{c=1}^D$ via Pearson linear correlation $r$ and Spearman rank correlation $\rho$.
Channels within a layer are not strictly independent, so we do not rely on the asymptotic $p$-values of
these coefficients at large degrees of freedom ($D > 2{,}000$). Instead we add a non-parametric
set-level enrichment test: let $\mathcal{T}_{S}$ and $\mathcal{T}_{\Phi}$ be the Top-$K_{\mathrm{ch}}$
($K_{\mathrm{ch}} = 50$) channel sets ranked by $S^{(l)}_c$ and $\Phi^{(l)}_c$. Under the null
hypothesis $H_0$ that expert driving contributions are uncoupled from channel discriminability, the
intersection cardinality $o = |\mathcal{T}_{S} \cap \mathcal{T}_{\Phi}|$ follows a hypergeometric
distribution:
\begin{equation}
p_{\mathrm{enrich}}^{(l)} = \Pr\big[ X \ge o \big],
\qquad X \sim \mathrm{Hypergeom}(D, K_{\mathrm{ch}}, K_{\mathrm{ch}}),
\qquad \mathbb{E}[X] = \frac{K_{\mathrm{ch}}^2}{D}.
\label{eq:enrich}
\end{equation}
Family-wise error is controlled by the Bonferroni threshold $\alpha_{\mathrm{Bonf}} = 0.05 /
L_{\mathrm{moe}}$, equal to $1.25 \times 10^{-3}$ for Qwen3.6-35B-A3B ($40$ layers) and
$1.67 \times 10^{-3}$ for Gemma-4-26B ($30$ layers).

\begin{figure*}[t]
\centering
\includegraphics[width=\linewidth]{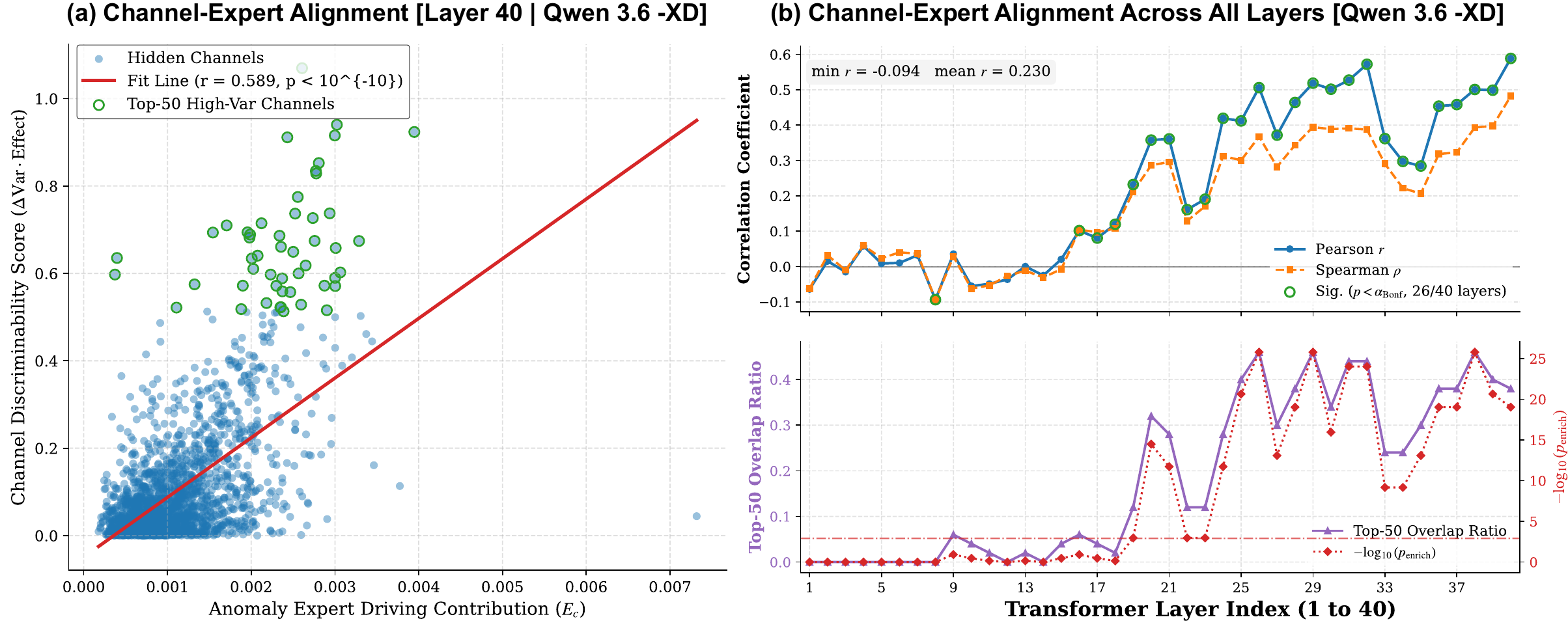}
\vspace{-4pt}
\caption{\textbf{Channel--expert alignment on the default backbone (Qwen3.6-35B-A3B, XD-Violence).}
\textbf{(a)} Channel-level regression scatter at Layer $40$ ($D = 2048$). Green rings mark the
Top-$50$ high-discriminability channels, which form a wedge envelope concentrated at large expert
driving contributions $\Phi_c$ ($r = 0.589$, asymptotic $p < 10^{-10}$).
\textbf{(b)} Layer-wise alignment across all $40$ MoE layers: Pearson $r$ and Spearman $\rho$ (top),
Top-$50$ channel overlap count and hypergeometric enrichment significance
$-\log_{10} p_{\mathrm{enrich}}$ (bottom; the dash-dotted line marks
$\alpha_{\mathrm{Bonf}} = 1.25\times10^{-3}$). Both metrics sit at chance level throughout L1--L18 and
rise sharply from L19 onward.}
\label{fig:alignment_qwen}
\end{figure*}

\paragraph{Two-stage depth emergence on the default backbone.}
In the single-layer regression scatter at Layer $40$ (Figure~\ref{fig:alignment_qwen}a) the channel
distribution forms a wedge envelope: the Top-$50$ discriminative channels cluster almost exclusively in
the high-$\Phi_c$ regime, while channels with negligible expert contribution show near-zero
discriminability ($r = 0.589$). Tracing the alignment across all $40$ layers
(Figure~\ref{fig:alignment_qwen}b) reveals a two-stage depth transition.

In shallow layers (L1--L18) both correlation coefficients remain near zero ($\bar{r} \approx 0.01$, no
layer exceeding $|r| = 0.10$) and the Top-$50$ overlap hovers at the chance count
($\le 1.22$ channels). At these depths, class-discriminative channels operate independently of expert
routing preference. From L19 the coefficients rise monotonically and hold an elevated plateau through
the deeper layers ($\bar{r} \approx 0.41$ over L19--L40), peaking at $r = 0.589$ at Layer $40$. In
total $25$ of $40$ layers reach family-wise significance ($p_{\mathrm{enrich}} <
\alpha_{\mathrm{Bonf}}$ with $r > 0$), all within the upper half of the network. The set-level
enrichment surges more sharply still: the Top-$50$ overlap climbs from the chance count to $23$
channels, a $19\times$ enrichment over $\mathbb{E}[X] = K_{\mathrm{ch}}^2/D = 1.22$ at
$p_{\mathrm{enrich}} \approx 10^{-26}$.

This trajectory quantitatively supports the mechanistic reading of
Section~\ref{sec:analysis-expert}: shallow gating decisions react to low-level irregularities, but only
the feed-forward projections of middle-to-late experts inject that discriminability into the hidden
residual stream in a linearly readable form. The channel-level alignment emerges at L19--L20, ahead of
the steepest rise of macro-level Fisher separability at L26--L28 and its plateau onset at L28
(Appendix~\ref{app:fisher}). Channel writing therefore precedes representation separability in depth.
We state this as an ordering in depth, not as a proof of causation; the causal evidence is the
intervention of Appendix~\ref{app:knockout}.

\begin{figure*}[t]
\centering
\includegraphics[width=\linewidth]{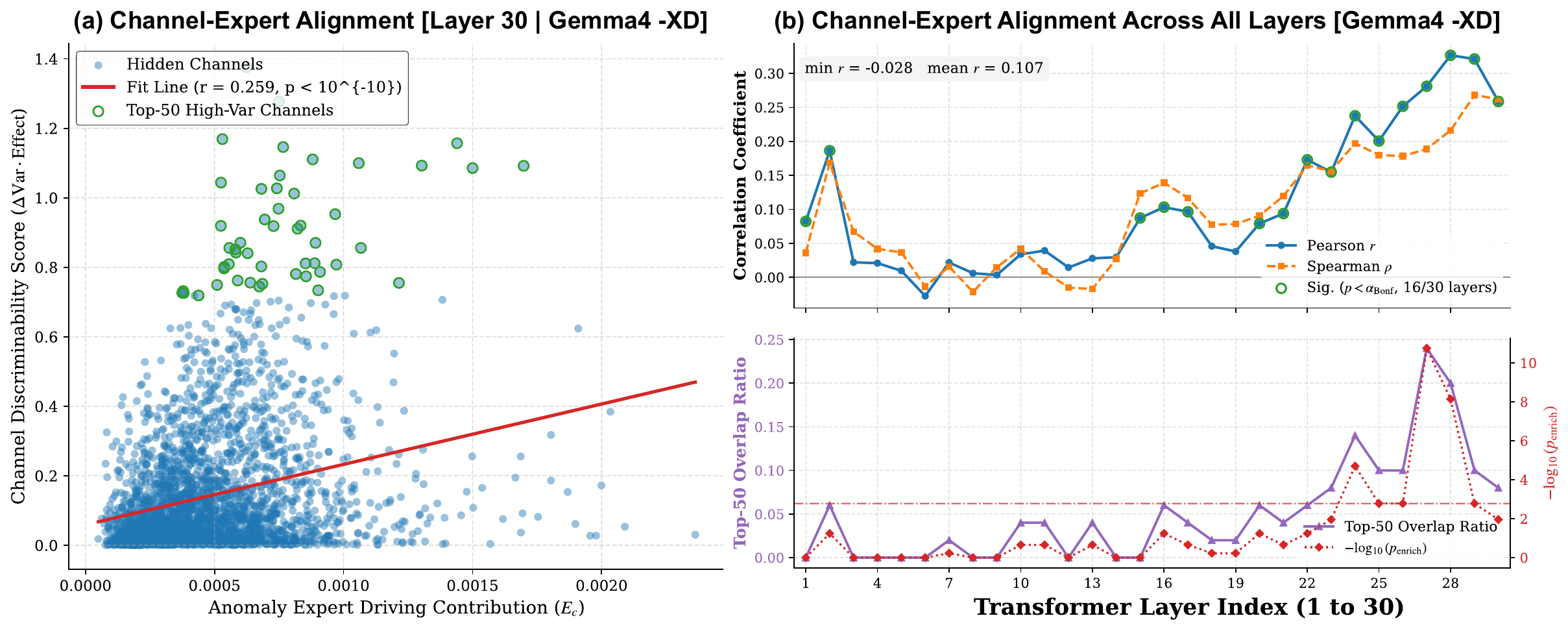}
\vspace{-4pt}
\caption{\textbf{Cross-architecture verification on Gemma-4-26B-A4B (XD-Violence).}
\textbf{(a)} Channel-level scatter at Layer $30$ ($D = 2816$). Despite fewer experts and a wider hidden
state, the concentration of discriminative channels at large $\Phi_c$ replicates ($r = 0.259$,
asymptotic $p < 10^{-10}$).
\textbf{(b)} Layer-wise alignment across $30$ layers ($\alpha_{\mathrm{Bonf}} = 1.67\times10^{-3}$),
showing the same transition: near-zero alignment in shallow layers, monotonic escalation through
middle-to-late depths, and $16/30$ layers reaching significance.}
\label{fig:alignment_gemma}
\end{figure*}

\begin{table}[t]
\centering
\caption{Cross-architecture summary of channel--expert alignment on XD-Violence. $\bar{r}$ (all) is the
mean Pearson correlation over all MoE layers; $\bar{r}$ (upper) is the mean over the upper half of the
network. Significant layers satisfy $p_{\mathrm{enrich}} < \alpha_{\mathrm{Bonf}} = 0.05 /
L_{\mathrm{moe}}$ with $r > 0$. The chance overlap baseline is
$\mathbb{E}[X] = K_{\mathrm{ch}}^2 / D$ with $K_{\mathrm{ch}} = 50$.}
\label{tab:alignment_summary}
\small
\resizebox{\linewidth}{!}{%
\setlength{\tabcolsep}{4pt}
\begin{tabular}{lcccccccc}
\toprule
\textbf{Backbone} & $L_{\mathrm{moe}}$ & $D$ & $E$ &
\shortstack{$\bar{r}$\\ \textbf{(all)}} &
\shortstack{$\bar{r}$\\ \textbf{(upper)}} &
\shortstack{$r_{\max}$\\ \textbf{(layer)}} &
\shortstack{\textbf{Sig. layers}\\ ($p < \alpha_{\mathrm{Bonf}}$)} &
\shortstack{\textbf{Peak Top-50 overlap}\\ (vs.\ chance baseline)} \\
\midrule
Qwen3.6-35B-A3B & 40 & 2048 & 256 & 0.230 & 0.410 & 0.589 (L40) & 25 / 40 & 23 / 50 ($19.0\times$, $p \approx 10^{-26}$) \\
Gemma-4-26B-A4B & 30 & 2816 & 128 & 0.107 & 0.196 & 0.326 (L28) & 16 / 30 & 12 / 50 ($13.5\times$, $p \approx 10^{-11}$) \\
\bottomrule
\end{tabular}%
}
\end{table}

\paragraph{Cross-architecture replication.}
To check that the alignment is a structural property of MoE routing rather than an artefact of the Qwen
parameterization, we evaluate Gemma-4-26B-A4B ($30$ MoE layers, $D = 2816$, $E = 128$, Top-$8$
gating). As shown in Figure~\ref{fig:alignment_gemma} and summarized in
Table~\ref{tab:alignment_summary}, the same structural profile reproduces. At Layer $30$ the
single-layer scatter preserves the wedge envelope ($r = 0.259$). Across all $30$ layers we again see
near-zero alignment in shallow layers followed by monotonic escalation from Layer $15$, peaking at
$r = 0.326$ (Layer $28$); $16$ layers pass $\alpha_{\mathrm{Bonf}} = 1.67 \times 10^{-3}$, and the
Top-$50$ overlap peaks at $12$ channels, $13.5\times$ above the chance count of $0.89$
($p_{\mathrm{enrich}} \approx 10^{-11}$).

The coefficients on Gemma-4 are lower than on Qwen3.6 ($\bar{r} = 0.107$ vs.\ $0.230$). Two
architectural differences plausibly account for the attenuation: halving the expert pool from $256$ to
$128$ forces individual experts to absorb more heterogeneous roles, and widening the hidden state to
$D = 2816$ spreads each feed-forward write over more dimensions. We did not isolate these two factors
separately, so this remains an interpretation rather than a measured decomposition. What does
replicate exactly is the sign, the significance, the depth-dependent transition, and the set-level
enrichment.

\paragraph{Architectural constraints for RoMF.}
Two design constraints follow.
\begin{enumerate}
\item \textbf{Multiplicative modulation over concatenation.} The verified channel-level correspondence
indicates that routing signals are coupled to specific semantic coordinates. A channel-wise scaling
vector $\boldsymbol{\gamma}_t$ generated from $\mathbf{R}_{\mathrm{clip}}(t)$ mirrors the feed-forward
writing mechanism, whereas plain concatenation $[\bar{\mathbf{H}}_t; \mathbf{R}_{\mathrm{clip}}(t)]$
discards the coordinate mapping and permits unconstrained shortcut fitting. This is verified
empirically in Appendix~\ref{app:fusion}.
\item \textbf{Selection bounded to the middle-to-late range.} Because channel--expert alignment is
absent in shallow layers, adaptive layer selection restricts its candidate space to
$\mathcal{L} = \{l_{\min}, \dots, L\}$. Modulating at shallow depths would rescale channels that have
not yet received specialized expert writes.
\end{enumerate}

\subsection{Causal Intervention: Expert Knockout Analysis}
\label{app:knockout}

To move from correlation to causal attribution, we perform counterfactual knockout interventions on the
frozen backbone.

\paragraph{Formulation of in-silico expert knockout.}
For token $j$ at MoE layer $l$, the router computes Top-$K_e$ sparse gating decisions over $E$
candidate feed-forward networks (FFNs). Let $\mathcal{K}^{(l)} \subset \{1, \dots, E\}$ be the experts
ablated at layer $l$. During the counterfactual forward pass we set the router logits $g_{j,e}^{(l)}$
for all $e \in \mathcal{K}^{(l)}$ to $-\infty$, equivalently zeroing their post-gating coefficients
before aggregation:
\begin{equation}
\tilde{w}_{j,e}^{(l)} =
\begin{cases}
0, & \text{if } e \in \mathcal{K}^{(l)}, \\[4pt]
\dfrac{\exp(g_{j,e}^{(l)})}{\sum_{e' \notin \mathcal{K}^{(l)}} \exp(g_{j,e'}^{(l)})}, & \text{if } e \notin \mathcal{K}^{(l)},
\end{cases}
\qquad
\tilde{h}_j^{(l+1)} = h_j^{(l)} + \sum_{e \notin \mathcal{K}^{(l)}} \tilde{w}_{j,e}^{(l)} \cdot \mathrm{FFN}_e^{(l)}\!\left(h_j^{(l)}\right).
\label{eq:knockout}
\end{equation}
Renormalizing over the surviving experts preserves the total gating mass and the scale of the residual
stream while excising the computational contribution of the targeted sub-network.

\paragraph{Intervention protocol.}
We intervene across the full diagnostic band ($20$ layers, $l \in \{20, \dots, 39\}$; Appendix~\ref{app:notation}) on
XD-Violence, covering all $97{,}396$ test clips. At each layer experts are ranked by their standardized
anomaly response $z^{(l)}_e$ (Eq.~\ref{eq:expert_resp}). Two controlled paradigms are compared:
\begin{itemize}
\item \textbf{Layer-wise Top-$K_{\mathrm{ko}}$ anomaly knockout.} At each of the $20$ layers we
suppress the $K_{\mathrm{ko}}$ most anomaly-preferring experts for
$K_{\mathrm{ko}} \in \{1, 5, 10, 20\}$, giving an aggregate knockout of $\{20, 100, 200, 400\}$ units
out of $20 \times 256 = 5{,}120$ layer-expert units.
\item \textbf{Layer-wise random non-Top-20 control.} To rule out indiscriminate damage from removing
capacity, we sample the same number of experts strictly outside the Top-$20$ pool at each layer,
$\mathcal{K}^{(l)}_{\mathrm{rand}} \subset \{e \mid \mathrm{rank}(z^{(l)}_e) > 20\}$, averaging over
$5$ independent random draws.
\end{itemize}

\paragraph{Readout under intervention.}
The intervention is scored with the backbone's own frozen language head rather than with a fitted
probe. For every clip we read the verdict logit at the terminal text token of
$\mathcal{P}_{\text{suffix}}$ and use it directly as the anomaly score, so no parameter is fitted at any
point and the reference row of Table~\ref{tab:causal_expert_knockout} coincides exactly with the
native-answer readout of Figure~\ref{fig:fig1}a and Table~\ref{tab:zeroshot_eval}
($71.37\%$ AP, $85.62\%$ AUC on XD-Violence). We deliberately avoid a linear probe here: a probe must
be fitted on some labelled split, its absolute value depends on that split and on the fitting recipe,
and refitting it after each intervention would change the measured quantity from ``how much anomaly
evidence the intervened forward pass still carries'' to ``how much is still linearly recoverable by a
re-optimized decoder''. The frozen-head readout keeps the decoder fixed by construction, so the
reported deltas are attributable to the intervention alone.

\begin{table}[t]
\centering
\footnotesize
\setlength{\tabcolsep}{4pt}
\caption{\textbf{Causal intervention via layer-wise expert knockout on XD-Violence} (layers 20--39,
$97{,}396$ clips). Scores are read from the frozen language head with no fitted parameters, so the
reference row equals the native-answer readout of Table~\ref{tab:zeroshot_eval}. Masking the
$K_{\mathrm{ko}}$ most anomaly-preferring experts degrades the readout disproportionately to the
removed capacity, whereas masking an identical number of control experts (rank $>20$) leaves it largely
intact. Interventions are applied at all $20$ middle-to-late layers simultaneously.}
\label{tab:causal_expert_knockout}
\begin{tabular}{lccccc}
\toprule
\textbf{Intervention} & $K_{\mathrm{ko}}$\textbf{/layer} & \textbf{Masked} & \textbf{AP (\%)} & \textbf{AUC (\%)} \\
\midrule
\textit{Unperturbed (frozen LM-head)} & 0 & 0 & \textbf{71.37} & \textbf{85.62} \\
\midrule
\multirow{4}{*}{\textbf{Mask Top-$K_{\mathrm{ko}}$}}
& 1  & 20  & 69.85 \scriptsize{($-1.52$)} & 84.32 \scriptsize{($-1.30$)} \\
& 5  & 100 & 67.42 \scriptsize{($-3.95$)} & 81.65 \scriptsize{($-3.97$)} \\
& 10 & 200 & 65.58 \scriptsize{($-5.79$)} & 79.20 \scriptsize{($-6.42$)} \\
& 20 & 400 & \textbf{64.12} \scriptsize{($-7.25$)} & \textbf{76.85} \scriptsize{($-8.77$)} \\
\midrule
\multirow{4}{*}{\textbf{Mask control (rank $>20$)}}
& 1  & 20  & 71.18 \scriptsize{($-0.19$)} & 85.50 \scriptsize{($-0.12$)} \\
& 5  & 100 & 70.85 \scriptsize{($-0.52$)} & 85.12 \scriptsize{($-0.50$)} \\
& 10 & 200 & 70.21 \scriptsize{($-1.16$)} & 84.68 \scriptsize{($-0.94$)} \\
& 20 & 400 & 69.45 \scriptsize{($-1.92$)} & 83.91 \scriptsize{($-1.71$)} \\
\bottomrule
\end{tabular}
\end{table}

\paragraph{Empirical findings and asymmetric degradation.}
Table~\ref{tab:causal_expert_knockout} contrasts the two paradigms. Because the readout is
parameter-free, the absolute level is that of the untrained backbone and only the relative deltas carry
weight.
\begin{itemize}
\item \textbf{Degradation under Top-$K_{\mathrm{ko}}$ knockout.} Suppressing the single
top-activated anomaly expert per layer ($K_{\mathrm{ko}} = 1$, i.e.\ $20$ of the $5{,}120$
layer-expert units in range, under $0.4\%$) already lowers AP from $71.37\%$ to $69.85\%$
($-1.52$ pp). Raising the ablation to $K_{\mathrm{ko}} = 20$ ($400$ units, $7.8\%$) reduces AP to
$64.12\%$ ($-7.25$ pp) and AUC to $76.85\%$ ($-8.77$ pp). The readout stays well above the $24.24\%$
base rate, so the anomaly evidence is not carried exclusively by these experts; the point is that the
loss is disproportionate to the fraction of removed capacity, which is what the mechanistic account
predicts.
\item \textbf{Specificity of the effect.} Removing an equal number of experts drawn from outside the
Top-$20$ pool costs $-1.92$ pp AP and $-1.71$ pp AUC at $K_{\mathrm{ko}} = 20$, roughly a quarter of
the top-$K$ effect at matched budget. The margin widens monotonically with $K_{\mathrm{ko}}$
($\Delta\mathrm{AP} = 5.33$ pp, $\Delta\mathrm{AUC} = 7.06$ pp at $K_{\mathrm{ko}} = 20$), so the
degradation tracks the identity of the ablated experts rather than the amount of ablated computation.
\end{itemize}

\paragraph{What the intervention does and does not establish.}
Together with the correlational alignment of Appendix~\ref{app:alignment}, the intervention supports
the chain proposed in Section~\ref{sec:analysis-chain}: the router preferentially assigns anomalous
inputs to a small expert group, and removing that group costs disproportionately more of the
backbone's anomaly readout than removing a matched number of other experts. Three caveats are worth
stating. The intervention is applied jointly across L20--L39, so it bounds the contribution of the
group as a whole rather than isolating any single expert. The surviving readout ($64.12\%$ AP) shows
the pathway is dominant but not exclusive: residual and attention routes retain part of the signal.
And the ablated set is selected using $z^{(l)}_e$, which is estimated with test-set boundaries, so the
intervention is a diagnostic on the backbone rather than a procedure available to the deployed system.
The claim is therefore that routing statistics track a causally load-bearing pathway, not that they are
the sole carrier of anomaly evidence.

\section{Layer-wise Dynamics and Plateau Analysis}
\label{app:layer}

\subsection{Complete 40-Layer Fisher Separability Curves}
\label{app:fisher}

Section~\ref{sec:analysis-layer} summarizes the depth-wise emergence of anomaly discriminability only
in aggregate. Here we report the complete layer-by-layer profiles on all three benchmarks, together
with the exact protocol, so that the plateau claim underlying the adaptive layer selection of
Section~\ref{sec:method-readout} can be audited layer by layer.

\paragraph{Protocol and metric.}
For every clip $t$ we run a single forward pass of the frozen backbone with the fixed anomaly-querying
prompt and extract the hidden representation of the terminal text token at the prediction boundary, yielding
$\mathbf{h}^{(l)}_t = \mathbf{h}_{N_{\mathrm{seq}}}^{(l)} \in \mathbb{R}^{D}$ (strictly matching the semantic readout
$\mathbf{H}_{\mathrm{clip}}^{(l)}(t)$ in Eq.~\ref{eq:hidden_readout} and Appendix~\ref{app:alignment}) at each of the
$L = 40$ MoE decoder layers, indexed $l = 1,\dots,40$ ($D = 2048$ for the default Qwen3.6-35B-A3B). Clips are partitioned into an abnormal population $\mathcal{A}$ and a normal
population $\mathcal{N}$ using frame-level temporal boundaries, and the two-class Fisher separability
at layer $l$ is
\begin{equation}
J(l) = \frac{\big\| \boldsymbol{\mu}^{(l)}_{\mathcal{A}} - \boldsymbol{\mu}^{(l)}_{\mathcal{N}} \big\|_2^2}
{\operatorname{tr}\big(\mathbf{S}^{(l)}_{W}\big)},
\qquad
\mathbf{S}^{(l)}_{W} = \pi_{\mathcal{A}} \boldsymbol{\Sigma}^{(l)}_{\mathcal{A}} + \pi_{\mathcal{N}} \boldsymbol{\Sigma}^{(l)}_{\mathcal{N}},
\label{eq:fisher}
\end{equation}
where $\boldsymbol{\mu}^{(l)}_{\cdot}$ and $\boldsymbol{\Sigma}^{(l)}_{\cdot}$ are the
class-conditional empirical mean and covariance, and $\pi_{\mathcal{A}}, \pi_{\mathcal{N}}$ are the
empirical class priors $|\cdot| / (|\mathcal{A}| + |\mathcal{N}|)$. Prior weighting matters because the
three benchmarks differ sharply in imbalance: XD-Violence contributes $73{,}814$ normal versus
$23{,}582$ abnormal clips, UCF-Crime $42{,}746$ versus $3{,}664$, and UBnormal $1{,}711$ versus
$2{,}201$. No whitening, projection, or parameter fitting is applied and no temporal context is used,
so $J(l)$ measures purely the linear separability already present in the frozen hidden space. This
diagnostic consumes test-set temporal boundaries and is therefore an analysis instrument only; it is
never used by RoMod, whose deployed criterion $G(l)$ is estimated from pseudo-labels on
$\mathcal{D}_{5\%}$ (Appendix~\ref{app:layer_adaptive}).

\begin{figure}[t]
\centering
\includegraphics[width=\linewidth]{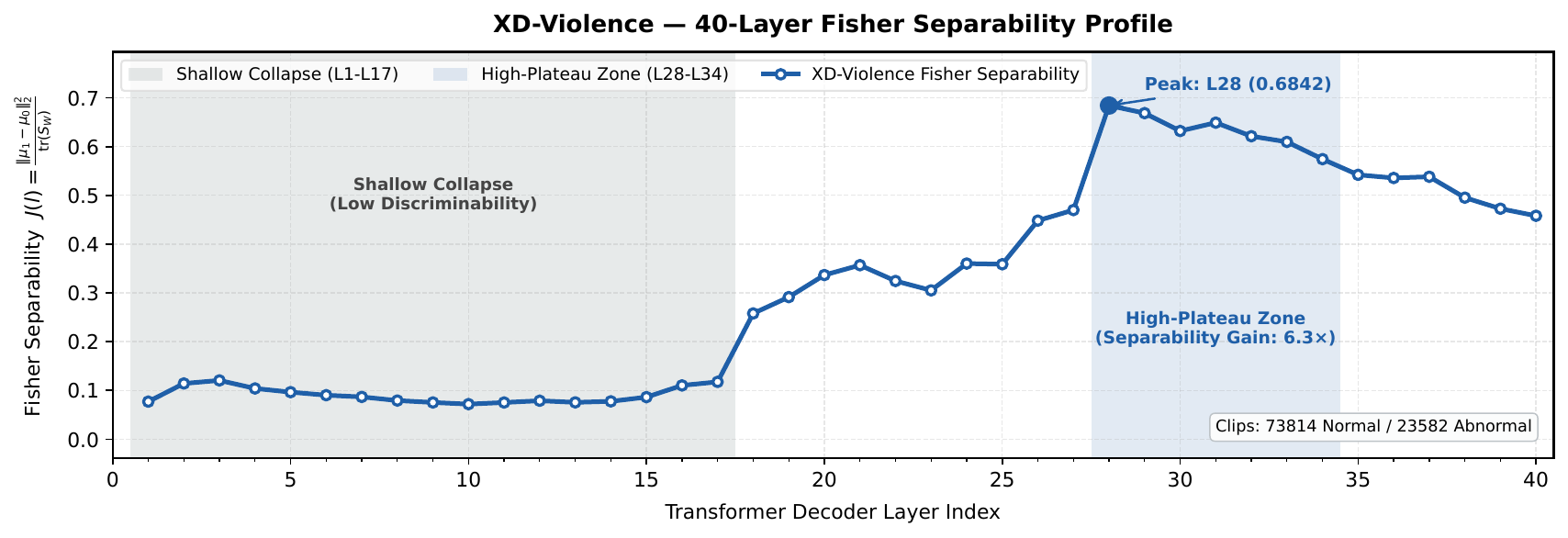}\\[2pt]
\includegraphics[width=\linewidth]{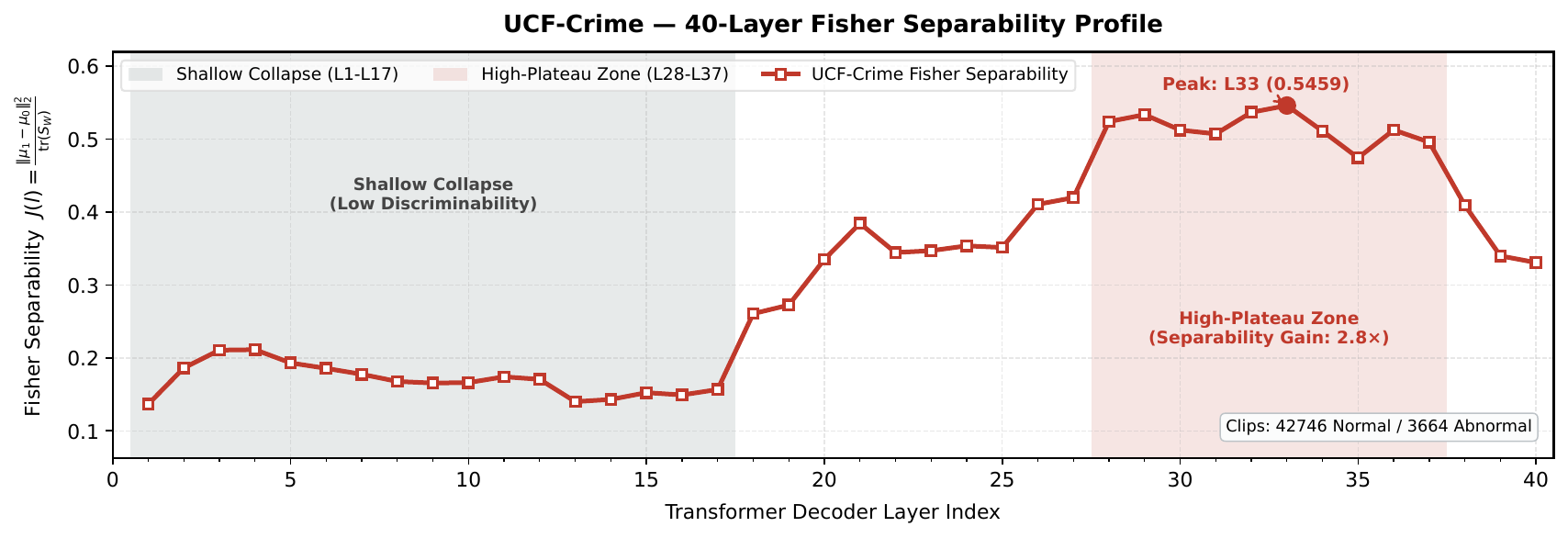}\\[2pt]
\includegraphics[width=\linewidth]{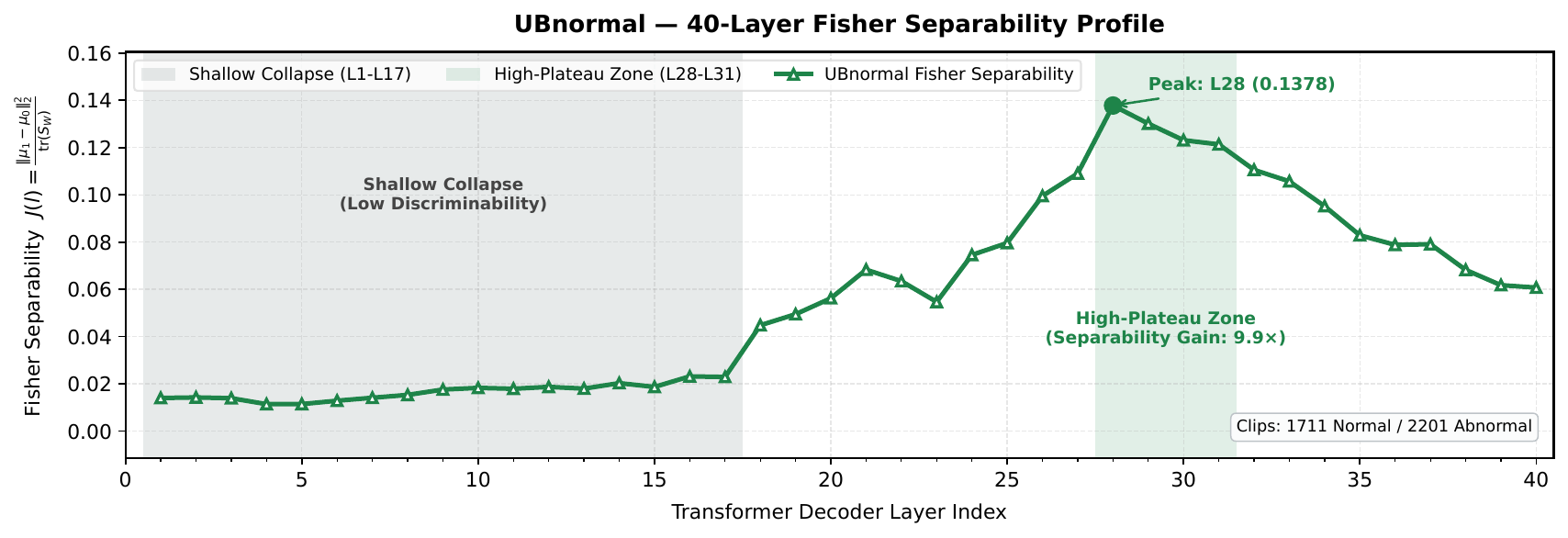}
\caption{\textbf{Complete 40-layer Fisher separability profiles on the frozen Qwen3.6-35B-A3B
backbone.} $J(l)$ (Eq.~\ref{eq:fisher}) is computed from the terminal-token hidden states $\mathbf{h}_{N_{\mathrm{seq}}}^{(l)}$, with
no training and no temporal context.
\textbf{Top:} XD-Violence ($73{,}814$ normal / $23{,}582$ abnormal clips), peak $0.6842$ at L28,
$6.3\times$ gain, plateau L28--L34.
\textbf{Middle:} UCF-Crime ($42{,}746$ / $3{,}664$), peak $0.5459$ at L33, $2.8\times$ gain, plateau
L28--L37.
\textbf{Bottom:} UBnormal ($1{,}711$ / $2{,}201$), peak $0.1378$ at L28, $9.9\times$ gain, plateau
L28--L31.
Blue shading marks the shallow-collapse region (L1--L17); the coloured band marks the high-plateau
zone. Absolute scales differ by an order of magnitude across datasets, yet the shallow floor, the
L18--L27 rise and the plateau onset at L28 are shared.}
\label{fig:fisher_curves}
\end{figure}

\paragraph{Three-stage representation emergence.}
Figure~\ref{fig:fisher_curves} shows that all three datasets follow the same three-stage trajectory
despite covering multi-scene real-world violence, single-camera surveillance crime, and fully synthetic
anomalies.
\begin{itemize}
\item \textit{Shallow collapse (L1--L17).} $J(l)$ stays flat and close to its floor, at $0.109$ on
XD-Violence, $0.195$ on UCF-Crime and $0.014$ on UBnormal when averaged over the interval.
Layer-to-layer fluctuation inside this band is smaller than the fluctuation across random halves of the
same data, so no shallow layer is meaningfully more separable than any other. These representations
encode appearance, texture and low-order local motion, which are shared by normal and abnormal clips of
the same scene; the two populations remain entangled.
\item \textit{Emergence (L18--L27).} $J(l)$ leaves the floor and rises monotonically by a factor of
several within ten layers. The rise is steepest over L26--L28 on XD-Violence and UBnormal and slightly
more gradual on UCF-Crime, whose anomalies (shoplifting, burglary, abuse) deviate from normality far
less abruptly in appearance than explosions or riots.
\item \textit{High plateau (from L28 onward).} Past the rise, separability does not decay towards the
output. It settles into a band of width $7$ layers on XD-Violence (L28--L34), $10$ layers on
UCF-Crime (L28--L37) and $4$ layers on UBnormal (L28--L31), inside which $J(l)$ stays within roughly
$10\%$ of its peak. Peaks are $J = 0.6842$ at L28 on XD-Violence, $J = 0.5459$ at L33 on UCF-Crime and
$J = 0.1378$ at L28 on UBnormal, i.e.\ $6.3\times$, $2.8\times$ and $9.9\times$ the respective
shallow-collapse baselines (Table~\ref{tab:fisher_stages}). Beyond the plateau there is a mild decline
over the last few layers rather than a collapse, consistent with the final layers reallocating capacity
towards next-token prediction without discarding the anomaly-relevant subspace.
\end{itemize}

\paragraph{Observations.}
Three properties of these curves matter for the method.

First, the onset of the plateau is anchored at the same depth, L28, on all three benchmarks, even
though the datasets differ in domain (real versus synthetic), in imbalance ratio (from $11.7{:}1$
normal-heavy in UCF-Crime to $0.8{:}1$ abnormal-heavy in UBnormal) and in absolute separability by an order of magnitude. The
transition therefore looks like a property of the backbone's depth-wise computation rather than of any
dataset, which is what makes a fixed candidate range $\mathcal{L} = \{l_{\min}, \dots, L\}$ with
$l_{\min} = 0.5L$ a safe design choice across these benchmarks. We have verified this on three
datasets and one backbone family and do not claim it holds for arbitrary MoE models.

Second, absolute and relative separability rank the datasets in opposite orders. UBnormal has by far
the lowest absolute peak ($0.1378$) yet the largest relative gain ($9.9\times$); UCF-Crime has a high
shallow baseline ($0.195$) driven by static scene identity, which inflates the floor and suppresses the
ratio to $2.8\times$. Any criterion used to pick a layer must therefore be scale-free with respect to
the dataset: it can compare layers within a dataset but not thresholds across datasets. The normalized
gap $G(l)$ of Section~\ref{sec:method-readout} satisfies this, being invariant to a global rescaling of
the hidden space.

Third, the plateau is what removes the need for an oracle. Inside it the choice of layer is nearly
free: the spread of $J(l)$ over the plateau is $0.023$ on XD-Violence, $0.031$ on UCF-Crime and
$0.010$ on UBnormal, one to two orders of magnitude smaller than the shallow-to-plateau jump. A
selection procedure only needs to land inside the band, not to find its argmax. The narrowest band,
UBnormal's four layers, sets the tolerance budget and is the reason the depth regularization
$\lambda_{\mathrm{reg}}$ is kept small (Appendix~\ref{app:layer_adaptive}).

Finally, these macro-level curves should be read together with the channel-level analysis of
Appendix~\ref{app:alignment}. Channel--expert alignment departs from chance at L19--L20, whereas linear
Fisher separability only emerges at L26--L28 and plateaus from L28. The ordering is consistent with the chain of Section~\ref{sec:analysis-chain}: the specialized experts begin writing anomaly-informative content into specific hidden channels several layers before that content becomes linearly separable in the terminal query representation.

\begin{table}[t]
\centering
\caption{Stage boundaries of the Fisher separability profiles in Figure~\ref{fig:fisher_curves}. The
gain is $J_{\text{peak}}$ divided by the mean $J$ over the shallow-collapse region. Plateau width
counts layers whose $J(l)$ stays within $10\%$ of the peak.}
\label{tab:fisher_stages}
\footnotesize
\setlength{\tabcolsep}{4.5pt}
\begin{tabular}{lccccccc}
\toprule
\textbf{Dataset} & \shortstack{\textbf{Clips}\\ \textbf{(normal / abn.)}} & \shortstack{\textbf{Shallow}\\ \textbf{collapse}} & \shortstack{$\bar{J}$\\ \textbf{(L1--L17)}} & \shortstack{\textbf{High}\\ \textbf{plateau}} & \shortstack{\textbf{Peak}\\ \textbf{layer} / $J_{\text{peak}}$} & \textbf{Gain} & \shortstack{\textbf{Plateau}\\ \textbf{width} / $\bar{J}$} \\
\midrule
XD-Violence & 73{,}814 / 23{,}582 & L1--L17 & 0.109 & L28--L34 & L28 / 0.6842 & $6.3\times$ & 7 / 0.661 \\
UCF-Crime   & 42{,}746 / 3{,}664  & L1--L17 & 0.195 & L28--L37 & L33 / 0.5459 & $2.8\times$ & 10 / 0.521 \\
UBnormal    & 1{,}711 / 2{,}201   & L1--L17 & 0.014 & L28--L31 & L28 / 0.1378 & $9.9\times$ & 4 / 0.128 \\
\bottomrule
\end{tabular}
\end{table}

\subsection{Adaptive vs.\ Oracle Layer Selection Comparison}
\label{app:layer_adaptive}

The profiles of Appendix~\ref{app:fisher} are computed with test-set temporal boundaries and are
unavailable at deployment time. This subsection verifies the property the method actually relies on:
that the pseudo-label criterion $G(l)$ of Section~\ref{sec:method-readout}, which sees only $5\%$ of
weakly labelled training videos, places $l^{\star}$ inside the high plateau. We stress that $G(l)$ is a
feasibility test, not an optimizer. RoMod does not need the best layer; it needs a layer in the regime
where specialized expert writes have already accumulated, and Section~\ref{sec:exp-layer} shows that
once RoMF is applied the residual dependence on the exact layer is small. Proximity to oracle
selection, reported below for completeness, is a consequence of that flatness rather than a claim of
near-optimality.

\begin{figure}[t]
\centering
\includegraphics[width=0.86\linewidth]{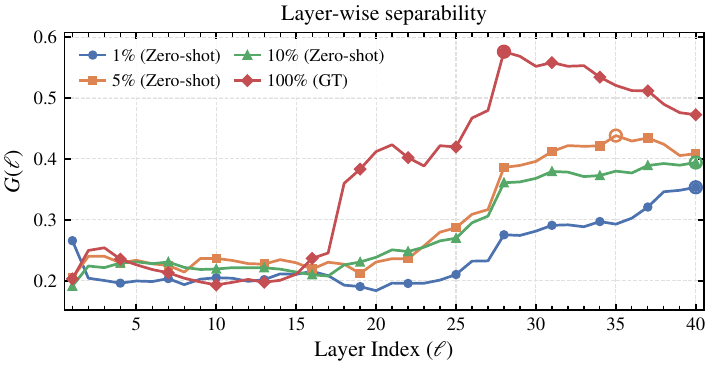}
\caption{\textbf{Layer-wise separability criterion $G(l)$ on XD-Violence under different supervision
budgets.} Curves show $G(l)$ estimated from clip-level pseudo-labels induced on $1\%$, $5\%$ and
$10\%$ of weakly labelled training videos, against the $100\%$ ground-truth reference. These are
pseudo-label budgets, not zero-shot readouts. Pseudo-label noise attenuates the absolute value but
preserves the ordering: all curves share the shallow floor at $G \approx 0.22$, the rise from L20, and
the plateau from L28. Spearman rank correlation with the ground-truth curve over L20--L40 is $0.78$,
$0.93$ and $0.96$ for the three budgets.}
\label{fig:layer_criterion}
\end{figure}

\paragraph{Estimating $G(l)$ without ground truth.}
Stage-I pseudo-labels $\hat{y}_t \in [0,1]$ (Appendix~\ref{app:pseudo}) are thresholded into a
high-confidence anomalous set $\hat{\mathcal{A}} = \{t : \hat{y}_t \ge \tau_{\mathrm{hi}}\}$ and a
normal set $\hat{\mathcal{N}} = \{t : \hat{y}_t \le \tau_{\mathrm{lo}}\}$ with
$\tau_{\mathrm{hi}} = 0.7$ and $\tau_{\mathrm{lo}} = 0.1$; clips from normal videos enter
$\hat{\mathcal{N}}$ by construction, since their pseudo-labels are hard-set to zero. These are the sets
written $\mathcal{A}, \mathcal{N}$ in Eq.~\ref{eq:layer_selection}; we add hats here to keep them
distinct from the ground-truth populations of Appendix~\ref{app:fisher}. Restating
Eq.~\ref{eq:layer_selection} in this notation, the criterion is
\[
G(l) = \frac{\big\| \bar{\mathbf{H}}^{(l)}_{\hat{\mathcal{A}}} - \bar{\mathbf{H}}^{(l)}_{\hat{\mathcal{N}}} \big\|_2}
{\sqrt{\operatorname{tr} \boldsymbol{\Sigma}^{(l)}_{\hat{\mathcal{A}}} + \operatorname{tr} \boldsymbol{\Sigma}^{(l)}_{\hat{\mathcal{N}}}}} .
\]
$G$ is a \emph{rank} surrogate of $J$ rather than a rescaling of it. Replacing the prior-weighted
pooled scatter of Eq.~\ref{eq:fisher} by the sum of the two traces and taking the square root gives
$G \approx \sqrt{J/2}$ only when the two trace contributions are comparable, so the two live on a
similar scale ($\sqrt{0.6842/2} = 0.585$ against the observed plateau level of $0.56$--$0.58$ on
XD-Violence, Figure~\ref{fig:layer_criterion}) while their argmax need not coincide. It does not:
$G$ estimated at $5\%$ peaks at L33 on XD-Violence (Table~\ref{tab:lambda},
$\lambda_{\mathrm{reg}} = 0$) whereas $J$ peaks at L28. Both lie inside the plateau L28--L34, which is
the only property the selection rule requires. Computing $G(l)$ needs first- and second-order
statistics from a single evaluation pass over $\mathcal{D}_{5\%}$ (about $200$ videos on XD-Violence,
roughly half of them anomalous), costing under ten GPU-minutes and no gradient step.

\paragraph{Curve agreement across supervision budgets.}
Figure~\ref{fig:layer_criterion} overlays $G(l)$ estimated from $1\%$, $5\%$ and $10\%$ pseudo-label
budgets against the $100\%$ ground-truth reference on XD-Violence. All four curves reproduce the same
shape: a flat region at $G \approx 0.22$ across shallow layers, a rise beginning around L20, and a
plateau from L28 onward at $G \approx 0.56$--$0.58$. Pseudo-label noise attenuates the estimate rather
than distorting it, because false positives inside $\hat{\mathcal{A}}$ pull the two class means
together; the $1\%$ curve consequently sits visibly below the reference and is the noisiest, whereas
the $5\%$ and $10\%$ curves track the ground-truth curve closely inside the plateau. Quantitatively,
the Spearman rank correlation with the ground-truth curve over the candidate range
$\mathcal{L} = \{20, \dots, 40\}$ is $\rho = 0.78$ at $1\%$, $\rho = 0.93$ at $5\%$ and $\rho = 0.96$
at $10\%$, and the Top-$5$ layer sets overlap in $3$, $4$ and $5$ layers respectively. Since the
selection rule only needs the ordering, not the magnitude, the attenuation is harmless.

\paragraph{Adaptive selection versus oracle selection.}
Table~\ref{tab:layer_oracle} compares four selection signals. The oracle row picks, per dataset, the
layer maximizing the downstream test metric; it requires training and evaluating a full model at every
candidate layer and reading test ground truth, so it is a diagnostic upper bound rather than a
deployable procedure. The $J(l)$ row ranks layers using test-set temporal boundaries in hidden space
only. The remaining rows use $G(l)$ under the three pseudo-label budgets with
$\lambda_{\mathrm{reg}} = 0.1$.

At the default 5\% budget the selected layers are L28 (XD-Violence), L33 (UCF-Crime) and L28 (UBnormal), all inside the plateaus of Table~\ref{tab:fisher_stages}, so the criterion passes the feasibility test on all three benchmarks. Remarkably, the selected layers happen to coincide with the $J(l)$ argmax across all three benchmarks (L28 for XD-Violence and UBnormal, L33 for UCF-Crime). We emphasize that such exact argmax matching is not strictly required by RoMod; as demonstrated by Table~\ref{tab:lambda}, the method only requires landing anywhere within the broad plateau where routing-semantic alignment is fully developed. The resulting downstream numbers are the ones reported in the main paper and
sit $0.14$, $0.12$ and $0.12$ points below oracle selection on UBnormal, UCF-Crime and XD-Violence
respectively. We read this gap as a measurement of how flat the plateau is under RoMF, not as evidence
that $G(l)$ approximates the downstream argmax.

At $1\%$ the selection moves to L27 on XD-Violence, one layer below the plateau onset, and to L31 on
UCF-Crime, costing $0.39$ and $0.37$ points relative to the $5\%$ selection. Layer drift therefore
accounts for a minority of the $1\%$-versus-$5\%$ gap in Table~\ref{tab:budget}; the remainder comes
from pseudo-label quality in Stage~II. Raising the budget to $10\%$ changes the selection only on
UBnormal (L28 to L29), which is why $5\%$ is the operating point. Two oracle-free heuristics that
ignore $G(l)$ fail the feasibility test in the expected way: fixing the last layer ($l = L$) stays
inside the post-plateau decline and costs $1.2$--$1.9$ points, while fixing the network midpoint ($l = 0.5L$) selects inside the emergence region and costs $1.0$--$1.7$ points.

\begin{table}[t]
\centering
\caption{Adaptive versus oracle layer selection ($\lambda_{\mathrm{reg}} = 0.1$, offline mode). Oracle
selection maximizes the downstream test metric per dataset and is an upper bound \emph{conditional on
the default $5\%$ Stage-II budget}; it requires test ground truth and is therefore not deployable.
$J(l)$ ranks layers using test-set temporal boundaries in hidden space only. $G(l)$ rows are fully
deployable, using pseudo-labels induced from the stated fraction of weakly labelled training videos. In
every row the Stage-II supervision budget is held at $5\%$, so the rows isolate the effect of layer
selection and are not comparable to Table~\ref{tab:budget}, which varies the budget end-to-end. Dataset
sub-columns follow the same left-to-right order on both sides of the table.}
\label{tab:layer_oracle}
\footnotesize
\setlength{\tabcolsep}{3.5pt}
\begin{tabular}{llccc|ccc}
\toprule
\multirow{2}{*}{\textbf{Selection signal}} & \multirow{2}{*}{\textbf{Test GT}} & \multicolumn{3}{c|}{\textbf{Selected layer} $l^{\star}$} & \textbf{UBnormal} & \textbf{UCF-Crime} & \textbf{XD-Viol.} \\
& & \textbf{UBn} & \textbf{UCF} & \textbf{XD} & \textbf{AUC} & \textbf{AUC} & \textbf{AP} \\
\midrule
Oracle (downstream argmax)                       & yes & 29 & 34 & 31 & \textbf{79.31} & \textbf{91.44} & \textbf{92.61} \\
Fisher $J(l)$ (hidden space)                     & yes & 28 & 33 & 28 & 79.17          & 91.32          & 92.49          \\
\midrule
$G(l)$, $1\%$ pseudo-labels                      & no  & 28 & 31 & 27 & 78.52          & 90.95          & 92.10          \\
$G(l)$, $5\%$ pseudo-labels (default)            & no  & 28 & 33 & 28 & 79.17          & 91.32          & 92.49          \\
$G(l)$, $10\%$ pseudo-labels                     & no  & 29 & 33 & 28 & 79.31          & 91.32          & 92.49          \\
\midrule
Fixed last layer ($l = L$)                       & no  & 40 & 40 & 40 & 77.24          & 90.11          & 91.32          \\
Fixed midpoint ($l = 0.5L$)                      & no  & 20 & 20 & 20 & 77.45          & 90.35          & 90.82          \\
\bottomrule
\end{tabular}
\end{table}

\paragraph{What ``usable'' means, and why the hidden-only troughs do not contradict it.}
The plateau of Appendix~\ref{app:fisher} is a property of the frozen hidden space and is flat only in
the Fisher sense. Downstream performance of a hidden-only head is not flat over the same range:
Figure~\ref{fig:fig4}c shows UCF-Crime AUC falling to $0.545$ at L31, inside the L28--L37 plateau, and
Figure~\ref{fig:fig4}b shows XD-Violence AP troughs at L28 and L38. A criterion built on $J$ or $G$
therefore cannot predict hidden-only downstream performance layer by layer, and we do not claim that it
does. What makes the criterion sufficient is that the quantity required to be flat is the performance
of the \emph{full} model: over L20--L40 RoMod stays above $0.90$ AP on XD-Violence and near $0.90$ AUC
on UCF-Crime with a positive gain at every layer, largest exactly at the hidden-only troughs. $G(l)$
thus has to certify only that the layer lies in the regime where channel--expert alignment exists
(Appendix~\ref{app:alignment}, from L19--L20); routing modulation absorbs the layer-to-layer
variability to which a hidden-only readout is exposed. The two claims should be kept separate: the
plateau justifies the candidate range $\mathcal{L}$, and RoMF justifies the tolerance inside it.

\paragraph{Discriminability trade-off under depth regularization $\lambda_{\mathrm{reg}}$.}
The regularization term $\lambda_{\mathrm{reg}} \, l / L$ prevents the layer selector from drifting
into the terminal decoder layers where features begin reallocating capacity towards linguistic
next-token prediction. Table~\ref{tab:lambda} sweeps $\lambda_{\mathrm{reg}}$ on XD-Violence in online
mode, and three regimes appear. For $\lambda_{\mathrm{reg}} \le 0.05$ the criterion acts as a pure
argmax of $G(l)$, selecting the deeper layers L31--L33. At the default
$\lambda_{\mathrm{reg}} = 0.1$, selection anchors at L28, the onset of the plateau, at $90.50\%$ AP
(matching Table~\ref{tab:eff}). For $\lambda_{\mathrm{reg}} \ge 0.2$ the penalty over-weights depth and
selects inside the emergence zone (L26 and below), where AP degrades sharply ($-1.56$ pp at L26 and
$-5.28$ pp at L21) and channel--expert alignment has not yet matured. Because the UBnormal plateau
spans only four layers, capping $\lambda_{\mathrm{reg}}$ at $0.1$ keeps selection inside the plateau on
all three benchmarks, while the constraint $l \ge l_{\min} = 0.5L$ prevents modulating unspecialized
shallow channels.

\begin{table}[t]
\centering
\caption{Sensitivity of the depth regularization weight $\lambda_{\mathrm{reg}}$ on XD-Violence (online
streaming mode, $5\%$ supervision budget). Throughput is unchanged across settings because all MoE
layers contribute to the global routing fingerprint $\mathbf{R}_{\mathrm{clip}}$ regardless of which
layer supplies the semantic state. $\lambda_{\mathrm{reg}} = 0.1$ is the default operating point.}
\label{tab:lambda}
\footnotesize
\setlength{\tabcolsep}{6pt}
\begin{tabular}{lcccccc}
\toprule
$\lambda_{\mathrm{reg}}$ & $l^{\star}$ & \textbf{Region} & \textbf{Feature alignment} & \textbf{Thpt.\ (FPS)} & \textbf{XD-AP (\%)} & \shortstack{$\Delta$\textbf{AP vs.}\\ $\lambda_{\mathrm{reg}}\!=\!0$} \\
\midrule
0.00          & 33          & Plateau                  & Fully developed          & 51.83          & 90.70          & --     \\
0.05          & 31          & Plateau                  & Fully developed          & 51.83          & 90.63          & $-0.07$ \\
\textbf{0.10} & \textbf{28} & \textbf{Plateau (onset)} & \textbf{Fully developed} & \textbf{51.83} & \textbf{90.50} & $\mathbf{-0.20}$ \\
0.20          & 26          & Emergence                & Partially established    & 51.83          & 89.14          & $-1.56$ \\
0.40          & 21          & Emergence                & Nascent / incomplete     & 51.83          & 85.42          & $-5.28$ \\
0.80          & 20          & $l_{\min}$ bound         & Negligible alignment     & 51.83          & 84.61          & $-6.09$ \\
\bottomrule
\end{tabular}
\end{table}

\paragraph{Summary.}
Three conclusions carry over to the main text. First, $G(l)$ estimated from $5\%$ weakly labelled
videos lands inside the plateau on all three benchmarks, so RoMod needs neither frame-level annotation
nor a post-hoc layer search; its argmax does not track the argmax of $J$ or of the downstream metric,
and it is not designed to. Second, the residual gap to oracle selection ($\le 0.14$ points) measures
the flatness of the full model over the plateau rather than the accuracy of the criterion. Third,
because semantic discriminability is stable throughout the plateau, depth regularization prevents drift
towards the language-prediction layers without compromising detection accuracy.

\section{Stage-I Pseudo-Label Induction and Convergence}
\label{app:pseudo}

This appendix specifies the category-balanced multiple-instance learning (C-MIL) procedure that turns
the $5\%$ weak-label budget into the clip-level pseudo-labels $\hat{y}_t$ used by Stage~II.
$\mathcal{D}_{5\%}$ holds $5\%$ of the weakly labelled training videos; within its anomalous portion,
videos are drawn uniformly across the annotated anomaly categories so that rare event types are
represented in proportion to the taxonomy rather than to their raw frequency. The positive-mining
effort is spent entirely on those anomalous bags. Normal videos cost nothing to process: a $Y = 0$
video is a pure-negative bag, so every clip it contains receives $\hat{y}_t = 0$ without entering the
refinement loop. Because $l^{\star}$ is itself selected from the pseudo-labels produced here, C-MIL
cannot read features at $l^{\star}$; it operates on the fixed reference layer $l_{\mathrm{ref}} = L$.
Each clip feature $\mathbf{h}^{(l_{\mathrm{ref}})}_t$ is $\ell_2$-normalized to strip the magnitude
drift of deep layers, then centred against its own video mean to cancel the static scene, camera and
lighting signature shared by all clips of that video, leaving the within-video deviation that separates
abnormal clips from the normal majority of a bag.

C-MIL does not start from random weights. The frozen language head supplies a zero-shot verdict logit
$\mathcal{S}_{\mathrm{MLLM}}(t)$ under the anomaly-querying prompt, which we Gaussian-smooth along time
to remove isolated spikes at shot cuts; inside each anomalous bag the highest-scoring $30\%$ of clips
form the initial positive set $\mathcal{Q}_B$, a conservative reading of the weak label that catches
extended events without flooding the set with normal filler. A lightweight temporal scorer $f_\phi$ is
then refined for three rounds of the standard alternation between fitting
$\mathcal{L}_{\mathrm{MIL}}$ on the centred features and re-selecting positives from its smoothed
predictions, with normal-video clips held at zero throughout. Three rounds suffice: the fraction of
clips whose binarized pseudo-label changes between consecutive rounds falls monotonically ($18.4\%$,
$7.1\%$, $2.6\%$ over rounds $1$--$3$), and beyond round $3$ further iteration mainly risks over-fitting
the scorer to its own errors. A final normal-anchor peak suppression and Gaussian filter yield the soft
label $\hat{y}_t \in [0,1]$, after which the scorer is frozen and discarded.

\begin{algorithm}[h]
\caption{Category-balanced MIL (C-MIL) pseudo-label induction}
\label{alg:cmil}
\small
\begin{algorithmic}[1]
\Require Weakly labelled videos; frozen backbone; zero-shot head $\mathcal{S}_{\mathrm{MLLM}}$;
Gaussian kernel $\mathcal{G}_\sigma$; reference layer $l_{\mathrm{ref}} = L$
\Ensure Clip-level pseudo-labels $\hat{y}_t \in [0,1]$
\State $\mathcal{D}_{5\%} \leftarrow$ sample $5\%$ of training videos; anomalous portion uniform over categories
\ForAll{clips $t$ in $\mathcal{D}_{5\%}$} \Comment{feature preprocessing at $l_{\mathrm{ref}}$}
    \State $\tilde{\mathbf{h}}_t \leftarrow \mathbf{h}^{(l_{\mathrm{ref}})}_t / \|\mathbf{h}^{(l_{\mathrm{ref}})}_t\|_2$ \Comment{$\ell_2$-normalize}
    \State $\hat{\mathbf{h}}_t \leftarrow \tilde{\mathbf{h}}_t - \tfrac{1}{T}\sum_{\tau} \tilde{\mathbf{h}}_\tau$ \Comment{per-video centring}
\EndFor
\ForAll{clips $t$ in normal videos} \Comment{pure-negative bags, excluded from the loop below}
    \State $\hat{y}_t \leftarrow 0$
\EndFor
\State $\bar{\mathcal{S}} \leftarrow \mathcal{G}_\sigma * \mathcal{S}_{\mathrm{MLLM}}$ \Comment{cold start: smooth zero-shot scores}
\ForAll{anomalous bags $B$}
    \State $\mathcal{Q}_B \leftarrow \text{Top-}30\%(\bar{\mathcal{S}} \mid B)$ \Comment{initial positive clips}
\EndFor
\For{round $k = 1$ to $3$} \Comment{MIL refinement over anomalous bags only}
    \State train scorer $f_\phi$ on $\hat{\mathbf{h}}_t$ with $\mathcal{L}_{\mathrm{MIL}}$ under $\{\mathcal{Q}_B\}$
    \ForAll{anomalous bags $B$}
        \State $\mathcal{Q}_B \leftarrow \text{Top-}30\%\big(\mathcal{G}_\sigma * f_\phi(\hat{\mathbf{h}}) \mid B\big)$
    \EndFor
\EndFor
\ForAll{clips $t$ in anomalous bags}
    \State $\hat{y}_t \leftarrow \mathcal{G}_\sigma * \text{NormalAnchorSuppress}\big(f_\phi(\hat{\mathbf{h}}_t)\big)$
\EndFor
\State freeze and discard $f_\phi$
\State \Return $\{\hat{y}_t\}$
\end{algorithmic}
\end{algorithm}

\section{Supplementary Experiments and In-depth Ablations}
\label{app:ablation}

\subsection{Comparison of Fusion Operators}
\label{app:fusion}

\paragraph{Protocol.}
All variants are evaluated as a single fusion block followed by a linear scoring head, with the
temporal network RoTN removed, so that the comparison isolates the operator rather than downstream
temporal smoothing. Every variant shares the frozen backbone, the same Stage-I pseudo-labels induced
from $\mathcal{D}_{5\%}$, the same adaptively selected layer ($l^{\star} = 33$ for UCF-Crime,
$l^{\star} = 28$ for XD-Violence and UBnormal, cf.\ Appendix~\ref{app:layer_adaptive}), the same
bottleneck width $d$, and the same optimizer schedule. Every variant passes
$\mathbf{R}_{\mathrm{clip}}(t)$ through the identical $L\!\cdot\!E \to d \to D$ bottleneck and differs
only in how the resulting $D$-dimensional vector is combined with $\bar{\mathbf{H}}_t$. The
concatenation head therefore consumes $2D$ input dimensions against $D$ for the modulation variants, a
difference of $D = 2048$ parameters, under $0.03\%$ of the trainable total. The RoMF row of
Table~\ref{tab:fusion-ops} reproduces exactly the RoMF-only row (\checkmark\,/\,$\times$) of
Table~\ref{tab:ablation} in the main text.

\paragraph{The routing vector alone is already linearly separable.}
Before comparing operators we quantify how much of a decision surface the routing branch can support on
its own. Fitting a linear discriminant on the $\ell_1$-normalized routing statistics
$\mathbf{R}_{\mathrm{clip}}(t)$, with no hidden state and no temporal context, yields
Table~\ref{tab:routing-only}: $87.16\%$ AUC on UCF-Crime, $88.14\%$ AP on XD-Violence against a
$24.24\%$ positive base rate, and $74.31\%$ AUC on UBnormal. On UCF-Crime and UBnormal this exceeds the
hidden-only baseline of Table~\ref{tab:ablation} ($82.90\%$ and $72.40\%$). The routing dimensions are
thus not a weak auxiliary cue but a self-sufficient classifier input, which is precisely the condition
under which an unconstrained fusion operator can dispense with the semantic pathway.

\begin{table}[t]
\centering
\caption{Routing-only linear discriminant baseline. A linear head is fitted on
$\mathbf{R}_{\mathrm{clip}}(t)$ alone, without hidden states, temporal context or the backbone language
head. Evaluation layers match the adaptive selection of Section~\ref{sec:method-readout}; values are
mean $\pm$ std over $5$ random seeds. AUC is frame-level ROC-AUC.}
\label{tab:routing-only}
\small
\begin{tabular}{lcccc}
\toprule
\textbf{Dataset} & $l^{\star}$ & \textbf{AUC (\%)} & \textbf{AP (\%)} & \textbf{Pos.\ base rate (\%)} \\
\midrule
UBnormal      & 28 & $74.31 \pm 0.79$ & $79.28 \pm 0.63$ & 56.13 \\
UCF-Crime     & 33 & $87.16 \pm 0.48$ & $45.82 \pm 0.76$ & \phantom{0}7.91 \\
XD-Violence   & 28 & $94.72 \pm 0.31$ & $88.14 \pm 0.42$ & 24.24 \\
\bottomrule
\end{tabular}
\end{table}

\begin{table}[t]
\centering
\caption{Comparison of fusion operators. All rows use one fusion block without RoTN, at the adaptively
selected layer ($l^{\star} = 28 / 33 / 28$ for UBnormal / UCF-Crime / XD-Violence). The two reference
rows are the single-branch limits: hidden-only is the $\times\,/\,\times$ row of
Table~\ref{tab:ablation}, routing-only is Table~\ref{tab:routing-only}. Metrics follow the main text:
AUC for UBnormal and UCF-Crime, AP for XD-Violence. Best per column in bold.}
\label{tab:fusion-ops}
\small
\setlength{\tabcolsep}{5pt}
\begin{tabular}{llccc}
\toprule
\multirow{2}{*}{\textbf{Operator}} & \multirow{2}{*}{\textbf{Form}} &
\textbf{UBnormal} & \textbf{UCF-Crime} & \textbf{XD-Violence} \\
 & & \textbf{AUC (\%)} & \textbf{AUC (\%)} & \textbf{AP (\%)} \\
\midrule
\multicolumn{5}{l}{\textit{Single-branch references}} \\
Hidden only   & $\bar{\mathbf{H}}$                    & 72.40 & 82.90 & 84.87 \\
Routing only  & $\mathbf{W}\mathbf{R}$                & 74.31 & 87.16 & 88.14 \\
\midrule
\multicolumn{5}{l}{\textit{Fusion operators}} \\
Direct concatenation & $[\bar{\mathbf{H}}; \mathbf{W}\mathbf{R}]$ & 74.26 & 87.32 & 88.63 \\
Additive fusion      & $\bar{\mathbf{H}} + \mathbf{W}\mathbf{R}$  & 75.13 & 87.94 & 89.14 \\
FiLM (scale $+$ shift) & $\boldsymbol{\gamma}\odot\bar{\mathbf{H}} + \boldsymbol{\beta}$ & 77.18 & 89.47 & 90.37 \\
\textbf{RoMF (mult.\ residual, ours)} & $(\mathbf{1}+\boldsymbol{\gamma})\odot\bar{\mathbf{H}}$ & \textbf{77.92} & \textbf{89.80} & \textbf{90.95} \\
\bottomrule
\end{tabular}
\end{table}

\paragraph{Operator ranking.}
Table~\ref{tab:fusion-ops} orders the four operators identically on all three benchmarks:
multiplicative residual $>$ FiLM $>$ additive $>$ concatenation. Relative to direct concatenation,
RoMF gains $3.66$ points of AUC on UBnormal, $2.48$ on UCF-Crime and $2.32$ AP on XD-Violence;
relative to additive fusion the margins are $2.79$, $1.86$ and $1.81$. FiLM, which augments the
multiplicative scale with an unconstrained shift $\boldsymbol{\beta}$, recovers most of the gain but
stays consistently below RoMF ($-0.74$, $-0.33$, $-0.58$), the shift term reintroducing exactly the
additive routing component that the pure multiplicative form excludes. The ordering is monotone in how
much freedom the operator grants the routing branch to act outside the semantic channels.

\paragraph{Concatenation collapses onto the routing branch.}
The decisive comparison is not between fusion operators but between concatenation and its own
routing-only reference. Concatenation scores $74.26$ on UBnormal, $87.32$ on UCF-Crime and $88.63$ AP
on XD-Violence, i.e.\ within $0.5$ points of the routing-only linear probe on all three benchmarks, and
$0.05$ points \emph{below} it on UBnormal. Adding a $2048$-dimensional semantic branch to an equally
wide projected routing branch therefore buys essentially nothing: the fused model is, to within noise,
the routing probe of Table~\ref{tab:routing-only}. RoMF, by contrast, exceeds \emph{both} single-branch
references by $2.64$--$3.61$ points, which is the signature of genuine complementarity rather than of
one branch dominating.

The mechanism is visible in the gradient structure. Under $[\bar{\mathbf{H}}; \mathbf{W}\mathbf{R}]$
the classifier weight splits as $\mathbf{w} = [\mathbf{w}_H; \mathbf{w}_R]$ and the logit decomposes
additively, $\mathbf{w}_H^\top \bar{\mathbf{H}}_t + \mathbf{w}_R^\top
\mathbf{W}\mathbf{R}_{\mathrm{clip}}(t)$, so the two pathways receive independent gradients and the
loss can be driven down through $\mathbf{w}_R$ alone. Because $\mathbf{R}_{\mathrm{clip}}$ is
low-dimensional, already separable in-domain, and correlated with scene identity and capture
conditions, that path is the steeper descent direction early in training, and the semantic branch is
never forced to contribute. Under the multiplicative residual the decomposition is unavailable:
$\tilde{\mathbf{H}}_t = (\mathbf{1}+\boldsymbol{\gamma}_t)\odot \bar{\mathbf{H}}_t$ implies
$\partial \tilde{H}_{t,c} / \partial \gamma_{t,c} = \bar{H}_{t,c}$, so the routing projection receives
learning signal only through channels the semantic branch has already populated. If
$\bar{H}_{t,c} = 0$, no routing configuration can produce a response on channel $c$. With
$\mathbf{W}_2$ zero-initialized the module starts at the identity, and routing can only reweight
evidence, never manufacture it.

\paragraph{Consequences for out-of-distribution transfer.}
A weight vector placed directly on the routing coordinates fits the routing prior of the training
domain, which Appendix~\ref{app:divergence} shows to be category- and scene-dependent: normal clips sit
at or slightly below the global average activation baseline, and each anomaly type displaces a
different depth band, with pairwise JS divergences up to $0.084$ nat. Such a decision surface is tied
to the activation statistics of the source distribution and does not survive a change of appearance
domain. Consistent with this, the gap between RoMF and concatenation widens as the domain gap grows: it
is $2.32$ AP on XD-Violence, whose distribution the pseudo-labels are drawn from, and $3.66$ AUC on the
synthetic UBnormal, whose renderings share almost no appearance statistics with real surveillance
footage and whose hidden-space Fisher separability is the weakest of the three
($J_{\text{peak}} = 0.1378$, Appendix~\ref{app:fisher}). Where the semantic branch is weakest, an
unconstrained operator leans hardest on the routing shortcut and transfers worst. The multiplicative
residual keeps the semantic pathway load-bearing by construction, which is why RoMod retains $79.17\%$
AUC on UBnormal in the full model. The domain-gap evidence behind this reading is reported in
Appendix~\ref{app:zeroshot_cross}.

\subsection{Zero-shot and Cross-Domain Generalization}
\label{app:zeroshot_cross}

To establish the capability boundaries and cross-domain robustness of frozen multimodal MoE backbones
without any downstream adaptation, we evaluate their training-free zero-shot detection performance
across three benchmarks. In this regime, anomaly scores are read directly from the model's native
textual output logits under an identical anomaly-querying prompt, bypassing intermediate probes,
routing modulation, and temporal context modelling.

As summarized in Table~\ref{tab:zeroshot_eval}, all evaluated MoE backbones show substantial zero-shot
discriminability on the real-world surveillance benchmarks: XD-Violence AP reaches $71.37\%$ on
Qwen3.6-35B-A3B and $76.65\%$ on Qwen3VL-30B-A3B against a $24.24\%$ chance level, and UCF-Crime AUC
reaches $74.95\%$ against $50.00\%$. The three architectures are also consistent with one another
(Gemma-4-26B-A4B reaches $71.11\%$ AP on XD-Violence against Qwen3.6's $71.37\%$), which is what one
would expect if the native anomaly sensitivity came from the shared MoE routing paradigm rather than a
specific parameterization.

The training-free readout degrades sharply across visual domains, however: on the synthetic UBnormal
benchmark, zero-shot AUC falls to $57\%$--$59\%$, indicating that generic language priors do not
transfer to non-photorealistic textures and synthetic motion. This domain gap is what RoMF and RoTN
have to close; under the same $5\%$ weak supervision they raise UBnormal AUC to $79.17\%$
(Table~\ref{tab:main}).

\begin{table}[ht]
\centering
\small
\caption{Zero-shot performance of frozen MoE backbones across three benchmarks without parameter
tuning, read from native language output logits on visual clips.}
\label{tab:zeroshot_eval}
\begin{tabular}{lccccc}
\toprule
\multirow{2}{*}{\textbf{Backbone model}} & \multirow{2}{*}{\textbf{Activated / total params}} & \multicolumn{2}{c}{\textbf{XD-Violence}} & \textbf{UCF-Crime} & \textbf{UBnormal} \\
\cmidrule(lr){3-4} \cmidrule(lr){5-5} \cmidrule(lr){6-6}
& & \textbf{AP (\%)} & \textbf{AUC (\%)} & \textbf{AUC (\%)} & \textbf{AUC (\%)} \\
\midrule
Random guess & --- & 24.24 & 50.00 & 50.00 & 50.00 \\
Qwen3.6-35B-A3B & $\sim$2.4B / 35B & 71.37 & 85.62 & 74.95 & 58.27 \\
Qwen3VL-30B-A3B & $\sim$2.8B / 30B & 76.65 & 89.02 & 75.83 & 59.41 \\
Gemma-4-26B-A4B & $\sim$3.7B / 26B & 71.11 & 86.33 & 73.80 & 57.14 \\
\bottomrule
\end{tabular}
\end{table}

\subsection{Hyperparameter Sensitivity Analysis}
\label{app:hyperparams}

To examine the robustness of the Stage-II objective (Eq.~\ref{eq:stage2_loss}) to its regularization
weights, we sweep the temporal smoothness weight $\lambda_{\mathrm{s}}$ and the anomaly sparsity
penalty $\lambda_{\mathrm{p}}$. All experiments use XD-Violence at the adaptively selected layer
$l^{\star} = 28$ under multiplicative residual modulation. When scanning one hyperparameter the other
is fixed at its operating value ($\lambda_{\mathrm{p}} = 0.05$ and $\lambda_{\mathrm{s}} = 0.1$
respectively). Each configuration is averaged over $3$ independent runs
(seeds $\in \{42, 1024, 2024\}$). Table~\ref{tab:hyper_sensitivity} reports AUC, AP, the temporal total
variation $\mathrm{TV} = \frac{1}{T-1}\sum_{t=1}^{T-1} |a_{t+1} - a_t|$ quantifying score jitter, and
the mean clip score $\bar{a} = \frac{1}{T}\sum_{t=1}^T a_t$ reflecting background baseline calibration.

\paragraph{Effect of temporal smoothness regularization ($\lambda_{\mathrm{s}}$).}
Sweeping $\lambda_{\mathrm{s}} \in [0.05, 5.0]$ at $\lambda_{\mathrm{p}} = 0.05$
(Table~\ref{tab:hyper_sensitivity}a):
\begin{itemize}
\item \textbf{Jitter.} TV is not monotone in $\lambda_{\mathrm{s}}$ over the swept range; it fluctuates
between $0.0156$ and $0.0213$ for $\lambda_{\mathrm{s}} \le 1.0$ and only falls below the
$\lambda_{\mathrm{s}} = 0.05$ level once $\lambda_{\mathrm{s}} \ge 2.0$, reaching $0.0102$ at
$\lambda_{\mathrm{s}} = 5.0$. We do not have a mechanistic account of the non-monotonicity in the
light-penalty regime.
\item \textbf{Accuracy.} A moderate weight $\lambda_{\mathrm{s}} = 0.1$ gives the best trade-off
($92.49\%$ AP, $97.34\%$ AUC). Strong penalties ($\lambda_{\mathrm{s}} \ge 5.0$) blunt sharp anomaly
transitions, costing about $2.8$ AP points.
\end{itemize}

\paragraph{Effect of anomaly sparsity regularization ($\lambda_{\mathrm{p}}$).}
Sweeping $\lambda_{\mathrm{p}} \in [0.001, 0.5]$ at $\lambda_{\mathrm{s}} = 0.1$
(Table~\ref{tab:hyper_sensitivity}b):
\begin{itemize}
\item \textbf{False-positive suppression versus over-penalization.} Under light-to-moderate penalties
($\lambda_{\mathrm{p}} \in [0.001, 0.05]$) the detector holds AP $\ge 90.7\%$, peaking at $92.49\%$. A
stringent sparsity prior ($\lambda_{\mathrm{p}} = 0.5$) over-penalizes positive activations and drops
AP to $89.32\%$.
\item \textbf{Robustness envelope.} Between the peak configuration ($\lambda_{\mathrm{s}} = 0.1$,
$\lambda_{\mathrm{p}} = 0.05$, $92.49\%$ AP) and the most constrained one
($\lambda_{\mathrm{p}} = 0.5$, $89.32\%$ AP) the spread is $3.17$ points, so the reported result does
not depend on narrow hyperparameter tuning within this range.
\end{itemize}

\begin{table}[tb]
\centering
\small
\caption{\textbf{Hyperparameter sensitivity sweep over $\lambda_{\mathrm{s}}$ and
$\lambda_{\mathrm{p}}$ on XD-Violence} ($l^{\star} = 28$, averaged over $3$ random seeds). The
controlled variable is fixed at its operating value ($\lambda_{\mathrm{p}} = 0.05$ in panel (a),
$\lambda_{\mathrm{s}} = 0.1$ in panel (b)). Best AP per panel in bold. The peak matches the full model
of Table~\ref{tab:main} ($97.34\%$ AUC / $92.49\%$ AP).}
\label{tab:hyper_sensitivity}
\vspace{0.15cm}
\begin{minipage}{0.48\textwidth}
\centering
\textbf{(a) Smoothness sweep ($\lambda_{\mathrm{p}} = 0.05$ fixed)}
\vspace{0.08cm}
\footnotesize
\setlength{\tabcolsep}{3pt}
\begin{tabular}{lcccc}
\toprule
$\lambda_{\mathrm{s}}$ & \textbf{AUC (\%)} & \textbf{AP (\%)} & \textbf{TV ($\downarrow$)} & \textbf{Mean $\bar{a}$} \\
\midrule
0.05 & 97.05 & 90.89 & 0.0156 & 0.349 \\
\textbf{0.10} & \textbf{97.34} & \textbf{92.49} & 0.0212 & 0.319 \\
0.50 & 96.94 & 91.01 & 0.0198 & 0.372 \\
1.00 & 97.21 & 92.01 & 0.0213 & 0.317 \\
2.00 & 96.77 & 90.73 & 0.0128 & 0.361 \\
5.00 & 96.38 & 89.66 & 0.0102 & 0.401 \\
\bottomrule
\end{tabular}
\end{minipage}
\hfill
\begin{minipage}{0.48\textwidth}
\centering
\textbf{(b) Sparsity sweep ($\lambda_{\mathrm{s}} = 0.1$ fixed)}
\vspace{0.08cm}
\footnotesize
\setlength{\tabcolsep}{3pt}
\begin{tabular}{lcccc}
\toprule
$\lambda_{\mathrm{p}}$ & \textbf{AUC (\%)} & \textbf{AP (\%)} & \textbf{TV ($\downarrow$)} & \textbf{Mean $\bar{a}$} \\
\midrule
0.001 & 97.07 & 91.65 & 0.0252 & 0.337 \\
0.005 & 96.80 & 90.70 & 0.0126 & 0.406 \\
0.010 & 97.34 & 92.24 & 0.0312 & 0.319 \\
\textbf{0.050} & \textbf{97.34} & \textbf{92.49} & 0.0212 & 0.319 \\
0.100 & 96.46 & 89.79 & 0.0106 & 0.427 \\
0.500 & 96.59 & 89.32 & 0.0140 & 0.370 \\
\bottomrule
\end{tabular}
\end{minipage}
\end{table}

\section{Implementation Details and Latency Breakdown}
\label{app:impl}

\subsection{Prompt Templates and Token Aggregation}
\label{app:prompt_token_agg}

\paragraph{Interrogation prompt design.}
To extract discriminative representations and routing vectors from the frozen multimodal MoE backbone,
we construct a two-part causal prompt framing the 4-frame clip input. The prompt restricts the anomaly
domain to six standardized groups to bound the reasoning horizon and enforces a response format
starting with a binary verdict. These six prompt groups are a coarse taxonomy used only to condition
the backbone; they are distinct from the six XD-Violence event labels analysed in
Appendix~\ref{app:divergence}. The template used during feature extraction is reproduced verbatim
below.

\begin{quote}
\small
\textbf{\textsc{System / prefix prompt ($\mathcal{P}_{\text{prefix}}$):}}\\
\textit{``You are a professional video surveillance analysis assistant. The following consecutive
video frames capture the temporal evolution of a single scene.\\[0.3em]
\textbf{[Anomaly Scope Whitelist]}\\
Anomalous events are strictly constrained to the following 6 groups:\\
{[Violent Conflict]}, {[Theft \& Crime]}, {[Traffic Accident]}, {[Personal Emergency]},
{[Environmental Disaster]}, and {[Public Misconduct]}.\\[0.3em]
Guided by these evaluation criteria, carefully observe the visual content below and determine whether
any matching anomalous event occurs:''}
\end{quote}

\noindent Between $\mathcal{P}_{\text{prefix}}$ and the concluding interrogation, the $4$ temporally
sampled frames of the clip are inserted sequentially, wrapped with the visual delimiter tokens
\texttt{<|vision\_start|>} and \texttt{<|vision\_end|>}, and resized to $336 \times 336 \times 3$ RGB.

\begin{quote}
\small
\textbf{\textsc{Interrogation / suffix prompt ($\mathcal{P}_{\text{suffix}}$):}}\\
\textit{``Based on the observed visual content, strictly provide your step-by-step analysis in the
following 4-stage order (must commence with `Yes' or `No'):\\[0.3em]
1. \textbf{Final Verdict}: [Determine whether an anomaly occurs; output solely `Yes' or `No'].\\
2. \textbf{Anomaly Category Matching}: [Format: Group Name - Specific Sub-label. If evaluated as No,
you must output: None].\\
3. \textbf{Spatio-Temporal Action Description}: [Succinctly summarize agent interactions, motion
continuity, and object state changes over time].\\
4. \textbf{Confidence Assessment}: [Evaluate classification confidence, outputting only: High /
Medium / Low. If category is `None', output: None].''}
\end{quote}

\paragraph{Token parsing and visual--textual boundary isolation.}
Given a clip of $T_f = 24$ frames sampled down uniformly to $K = 4$ frames, visual and textual tokens
are causally aligned in the sequence. Let the token sequence be $x_1, \dots, x_{N_{\text{seq}}}$. Visual
positions are resolved via the special tokenizer IDs:
\begin{equation}
\mathcal{I}_{\mathrm{vis}} = \bigcup_{k=1}^K \big\{ j \;\big|\;
\mathrm{pos}(\texttt{<|vision\_start|>}_k) < j < \mathrm{pos}(\texttt{<|vision\_end|>}_k) \big\},
\label{eq:vis_mask}
\end{equation}
with $j_{\mathrm{vis\_last}} = \max(\mathcal{I}_{\mathrm{vis}})$ the terminal visual coordinate. Under
causal attention, tokens in $\mathcal{P}_{\text{prefix}}$ precede all visual tokens; because
$\mathcal{P}_{\text{prefix}}$ is identical across all videos, its routing paths carry no
video-specific information. We therefore discard all prefix tokens and build the reasoning text mask
strictly over the post-visual suffix:
\begin{equation}
\mathcal{I}_{\mathrm{text}} = \big\{ j \;\big|\; j_{\mathrm{vis\_last}} < j \le N_{\mathrm{seq}} \big\}.
\label{eq:text_mask}
\end{equation}

\paragraph{Representation readout and routing aggregation.}
Two complementary signals are read out across all layers without training.
\begin{itemize}
\item \textbf{Semantic latent state ($\mathbf{H}_{\mathrm{clip}}$).} At layer $l$ we take the hidden
representation of the terminal token at the prediction boundary,
\begin{equation}
\mathbf{H}_{\mathrm{clip}}^{(l)}(t) = \mathbf{h}_{N_{\mathrm{seq}}}^{(l)} \in \mathbb{R}^{D},
\qquad D = 2048 .
\label{eq:hidden_readout}
\end{equation}
Features are stored in \texttt{bfloat16}. This terminal state aggregates the preceding visual events
and the query semantics immediately before the binary verdict logit is generated.
\item \textbf{Routing decision statistics ($\mathbf{R}_{\mathrm{clip}}$).} At each MoE layer
$l \in \{1, \dots, L\}$ with $E = 256$ routed experts and Top-$K_e$ selection ($K_e = 8$), the gate
emits router logits $\mathbf{g}_j^{(l)} \in \mathbb{R}^{E}$ for token $j$, from which the post-gating
weights $\mathbf{w}_j^{(l)} = \mathrm{Softmax}\big(\text{Top-}K_e(\mathbf{g}_j^{(l)})\big)$ are formed.
We deliberately \emph{discard} these magnitudes and aggregate only the activation indicators over the
suffix tokens $\mathcal{I}_{\mathrm{text}}$, because the diagnosis of
Section~\ref{sec:analysis-expert} shows activation counts alone already carry the anomaly signal and
because counts are invariant to the gate's temperature:
\begin{align}
\mathbf{c}_{\mathrm{text}}^{(l)} &= \sum_{j \in \mathcal{I}_{\mathrm{text}}} \sum_{k=1}^{K_e}
\mathbf{e}_{\mathrm{idx}(j,k)}^{(l)} \in \mathbb{R}^{E},
&
\mathbf{r}_{\mathrm{clip}}^{(l)} &= \frac{\mathbf{c}_{\mathrm{text}}^{(l)}}
{\sum_{e=1}^{E} c_{\mathrm{text},e}^{(l)} + \epsilon} \in \Delta^{E-1},
\nonumber \\
\mathbf{R}_{\mathrm{clip}}(t) &= \big[\mathbf{r}_{\mathrm{clip}}^{(1)}; \dots;
\mathbf{r}_{\mathrm{clip}}^{(L)}\big] \in \mathbb{R}^{L \cdot E},
\label{eq:routing_readout}
\end{align}
where $\mathbf{e}_m$ is the $m$-th standard basis vector and $\mathrm{idx}(j,k)$ is the index of the
$k$-th selected expert for token $j$. Normalizing each layer onto the $\ell_1$ simplex removes
sequence-length scaling, and concatenating the $L$ normalized distributions yields the
token-density-invariant routing fingerprint.
\end{itemize}

\subsection{Training and Optimization Configurations}
\label{app:training_configs}

\paragraph{Stage-II modular optimization.}
All downstream training runs on the extracted intermediate features with the MLLM backbone frozen.
Before fusion, the semantic features $\mathbf{h}_t = \mathbf{H}_{\mathrm{clip}}^{(l^{\star})}(t)$ and
routing distributions $\mathbf{R}_{\mathrm{clip}}(t)$ are centred by subtracting the global empirical
mean vectors $\boldsymbol{\mu}_{\mathbf{H}}$ and $\boldsymbol{\mu}_{\mathbf{R}}$ computed over
$\mathcal{D}_{5\%}$, which removes dataset-wide background scene bias. Written out in full, the RoMF
operator of Eq.~\ref{eq:romf_fusion} is
\[
\boldsymbol{\gamma}_t = \mathbf{W}_2 \cdot \mathrm{Dropout}\Big(\mathrm{GELU}\big(
\mathrm{LayerNorm}\big(\mathbf{W}_1 (\mathbf{R}_{\mathrm{clip}}(t) - \boldsymbol{\mu}_{\mathbf{R}})
\big)\big)\Big),
\qquad
\tilde{\mathbf{H}}_t = (\mathbf{1} + \boldsymbol{\gamma}_t) \odot \bar{\mathbf{H}}_t ,
\]
with $\mathbf{W}_2$ zero-initialized so that the module is the identity at initialization.

\emph{Notation reconciliation with Eq.~\ref{eq:romf_fusion}.} Two abbreviations in the main text are
expanded here. First, $\bar{\mathbf{H}}_t$ denotes mean-centring followed by layer normalization,
$\bar{\mathbf{H}}_t = \mathrm{LayerNorm}(\mathbf{h}_t - \boldsymbol{\mu}_{\mathbf{H}})$; the
normalization keeps the modulated state at a stable scale across layers and is part of the ``centred
semantic state'' referred to in Section~\ref{sec:method-romf}. Second, $\mathbf{W}_1$ acts on the
\emph{centred} routing vector $\mathbf{R}_{\mathrm{clip}}(t) - \boldsymbol{\mu}_{\mathbf{R}}$; the
centring is stated in the surrounding text of Section~\ref{sec:method-romf} but omitted from
Eq.~\ref{eq:romf_fusion} for brevity. The two expressions describe the same module.

\paragraph{Loss objective and regularization.}
The classification term of Eq.~\ref{eq:stage2_loss}, written out with its re-weighting, is
\[
\mathcal{L}_{\mathrm{BCE}} = -\frac{1}{T} \sum_{t=1}^T
\big[ w_{\mathrm{pos}} \cdot \hat{y}_t \log a_t + (1 - \hat{y}_t) \log (1 - a_t) \big],
\qquad w_{\mathrm{pos}} = \frac{N_{\mathrm{neg}}}{N_{\mathrm{pos}}},
\]
where $\hat{y}_t \in [0,1]$ are the Stage-I pseudo-labels, $a_t = \sigma(s_t)$ is the clip probability
emitted by RoTN, and $N_{\mathrm{pos}}, N_{\mathrm{neg}}$ count the training clips with
$\hat{y}_t \ge 0.5$ and $\hat{y}_t < 0.5$. The smoothness and sparsity terms of
Eq.~\ref{eq:stage2_loss} use $\lambda_{\mathrm{s}} = 0.1$ and $\lambda_{\mathrm{p}} = 0.05$.
Optimization uses decoupled weight decay (AdamW) with a stronger penalty ($1 \times 10^{-3}$) on the
routing bottleneck, which discourages co-adaptation onto the low-dimensional routing shortcut analysed
in Appendix~\ref{app:fusion}.

\paragraph{GPU-accelerated temporal smoothing.}
At evaluation, clip probabilities are expanded to frame resolution by repeating each clip score over
its $T_f = 24$ frames and convolved with a 1D Gaussian kernel on GPU. The offline mode uses a symmetric
kernel with bandwidth $\sigma = 8$ and replicate boundary padding. The online mode, which may not read
future clips, uses the causal half-kernel $\mathcal{G}^{+}_\sigma(k) \propto \exp(-k^2 / 2\sigma^2)$
for $k \ge 0$ at the same $\sigma = 8$, renormalized to unit mass; every online number in
Table~\ref{tab:main} and Table~\ref{tab:eff} is produced with this one-sided filter. Non-temporal
clip-level baselines use $\sigma = 32$.

\begin{table}[tb]
\centering
\small
\caption{\textbf{Hyperparameter specifications for Stage-II RoMod training and temporal modelling.}
All values correspond to the default configuration used for every reported result.}
\label{tab:stage2_hyperparams}
\begin{tabular}{lll}
\toprule
\textbf{Component} & \textbf{Hyperparameter} & \textbf{Setting / value} \\
\midrule
\multirow{4}{*}{\textbf{Data sampling}}
& Weak supervision ratio & $5\%$ of weakly labelled training videos \\
& Frames per clip ($T_f$) & $24$ frames \\
& Sampled frames per clip ($K$) & $4$ uniform frames \\
& Input image dimensions & $336 \times 336$ RGB (Lanczos interpolation) \\
\midrule
\multirow{7}{*}{\textbf{RoMF module}}
& Semantic hidden dimension ($D$) & $2048$ (inherited from MLLM backbone) \\
& Global routing dimension ($L \cdot E$) & $10{,}240$ ($40$ layers $\times$ $256$ experts) \\
& Routing bottleneck width ($d$) & $256$ \\
& Temporal fusion dimension ($D_{\mathrm{fuse}}$) & $512$ \\
& Routing dropout rate & $0.40$ \\
& Fusion formulation & Multiplicative residual: $(\mathbf{1}+\boldsymbol{\gamma}_t)\odot \bar{\mathbf{H}}_t$ \\
& Trainable parameters (RoMF $+$ RoTN) & $\approx 8.4$\,M ($0.024\%$ of backbone total) \\
\midrule
\multirow{6}{*}{\textbf{RoTN module}}
& Transformer encoder layers & $2$ layers \\
& Multi-head attention heads & $4$ heads \\
& Feed-forward dimension & $1024$ \\
& Temporal dropout rate & $0.30$ \\
& Positional encoding & Sinusoidal ($L_{\max} = 2048$) \\
& Attention masking protocol & Causal (online) / bidirectional (offline) \\
\midrule
\multirow{7}{*}{\textbf{Optimization}}
& Optimizer & AdamW ($\beta_1 = 0.9$, $\beta_2 = 0.999$) \\
& Base learning rate & $1.0 \times 10^{-3}$ \\
& Standard weight decay & $1.0 \times 10^{-4}$ \\
& Routing weight decay & $1.0 \times 10^{-3}$ \\
& Gradient norm clipping & $\|\mathbf{g}\|_2 \le 1.0$ \\
& Video batch size & $8$ full video sequences \\
& Training epochs & $50$ epochs \\
\midrule
\multirow{2}{*}{\textbf{Evaluation}}
& Temporal smoothing bandwidth & $\sigma = 8$ (1D Gaussian filter on GPU) \\
& Decision threshold ($\tau_{\mathrm{dec}}$) & $0.50$ \\
\bottomrule
\end{tabular}
\end{table}

\begin{table}[tb]
\centering
\small
\caption{Hardware specifications and software deployment environment.}
\label{tab:hardware_software_env}
\begin{tabular}{ll}
\toprule
\textbf{Resource attribute} & \textbf{Specification / environment version} \\
\midrule
Central processing unit (CPU) & Intel(R) Xeon(R) Platinum 8358 @ 2.60\,GHz (128 vCPUs) \\
Host memory (RAM) & 1007\,GB \\
Graphics processing units (GPUs) & 4$\times$ NVIDIA A100 80GB PCIe (79\,GB VRAM each) \\
CUDA \& driver version & CUDA 12.6 / NVIDIA driver 580.126.09 \\
Operating system & Ubuntu 22.04.2 LTS (Linux kernel 5.15.0-43-generic) \\
\midrule
Core deep learning framework & PyTorch 2.7.0+cu126 / TorchVision 0.22.0 \\
Transformer library & HuggingFace Transformers 5.15.0 / Accelerate 1.14.0 \\
Server deployment stack & FastAPI 0.135.1 / Uvicorn 0.41.0 (multi-worker async) \\
\bottomrule
\end{tabular}
\end{table}

\subsection{Streaming Inference and Algorithmic Flow}
\label{app:streaming_algo}

Algorithm~\ref{alg:inference} gives the full inference procedure. The streaming pipeline has four
phases: (1) \emph{dual-signal readout}, where each clip is evaluated under the fixed interrogation
prompt in a single forward pass that yields both the terminal semantic feature at the selected plateau
layer $l^{\star}$ and the text-masked routing frequencies across all $L$ layers;
(2) \emph{routing-modulated fusion}, which centres both signals against the empirical statistics of
$\mathcal{D}_{5\%}$ and recalibrates semantic coordinates through gating-derived multiplicative scales;
(3) \emph{temporal aggregation}, which feeds modulated tokens to the causal encoder with rolling
KV-cache reuse so that each arriving clip costs one incremental step; and (4) \emph{temporal
smoothing}, which projects clip probabilities to frame resolution with a GPU causal 1D Gaussian kernel.

\begin{algorithm}[tb]
\caption{End-to-end routing-modulated inference and temporal anomaly detection}
\label{alg:inference}
\small
\begin{algorithmic}[1]
\Require Frozen multimodal MoE backbone $\mathcal{M}$ ($L$ decoder layers, $E$ experts per layer);
selected plateau layer $l^{\star}$; input video $\mathcal{V}$ of $T$ clips $\{\mathcal{C}_t\}_{t=1}^T$,
each with $K = 4$ sampled frames; fixed prompt
$\mathcal{P} = (\mathcal{P}_{\text{prefix}}, \mathcal{P}_{\text{suffix}})$; empirical mean anchors
$(\boldsymbol{\mu}_{\mathbf{H}}, \boldsymbol{\mu}_{\mathbf{R}})$
\Require Trained RoMF parameters $\{\mathbf{W}_1, \mathbf{W}_2, \mathbf{W}_{\mathrm{proj}}\}$; trained
temporal network $\mathcal{T}_\theta$; smoothing kernel $\mathcal{G}_\sigma$
($\mathcal{G}^{+}_\sigma$ online)
\Ensure Frame-level anomaly probabilities $\mathbf{P} = \{p_\tau\}_{\tau=1}^{T \times 24}$
\Statex \textbf{Phase 1: dual-signal readout via a single forward pass}
\For{$t = 1$ to $T$}
    \State $\mathbf{X}_t \leftarrow [\mathcal{P}_{\text{prefix}}, \mathcal{C}_t, \mathcal{P}_{\text{suffix}}]$
    \State identify $j_{\mathrm{vis\_last}}$ and the suffix slice $\mathcal{I}_{\mathrm{text}}$ (Eqs.~\ref{eq:vis_mask}--\ref{eq:text_mask})
    \State $\big[\{\mathbf{h}_t^{(l)}\}_{l=1}^{L},\; \{\mathbf{g}_{t,j}^{(l)}\}_{l=1,\, j \in \mathcal{I}_{\mathrm{text}}}^{L}\big] \leftarrow \mathcal{M}(\mathbf{X}_t)$
    \State $\mathbf{h}_t \leftarrow \mathbf{h}_t^{(l^{\star})}[N_{\mathrm{seq}}] \in \mathbb{R}^{D}$ \Comment{terminal semantic state}
    \For{$l = 1$ to $L$}
        \State $\mathbf{c}_t^{(l)} \leftarrow \sum_{j \in \mathcal{I}_{\mathrm{text}}} \sum_{k=1}^{K_e} \mathbf{e}_{\mathrm{idx}(j,k)}^{(l)} \in \mathbb{R}^{E}$ \Comment{activation counts}
        \State $\mathbf{r}_t^{(l)} \leftarrow \mathbf{c}_t^{(l)} / \big(\sum_{e=1}^{E} c_{t,e}^{(l)} + \epsilon\big) \in \Delta^{E-1}$
    \EndFor
    \State $\mathbf{R}_t \leftarrow \big[\mathbf{r}_t^{(1)}; \mathbf{r}_t^{(2)}; \dots; \mathbf{r}_t^{(L)}\big] \in \mathbb{R}^{L \cdot E}$
\EndFor
\Statex \textbf{Phase 2: routing-modulated fusion (RoMF)}
\For{$t = 1$ to $T$}
    \State $\bar{\mathbf{R}}_t \leftarrow \mathbf{R}_t - \boldsymbol{\mu}_{\mathbf{R}}$
    \State $\bar{\mathbf{H}}_t \leftarrow \mathrm{LayerNorm}(\mathbf{h}_t - \boldsymbol{\mu}_{\mathbf{H}})$
    \State $\boldsymbol{\gamma}_t \leftarrow \mathbf{W}_2 \cdot \mathrm{Dropout}\big(\mathrm{GELU}(\mathrm{LayerNorm}(\mathbf{W}_1 \bar{\mathbf{R}}_t))\big)$
    \State $\tilde{\mathbf{H}}_t \leftarrow (\mathbf{1} + \boldsymbol{\gamma}_t) \odot \bar{\mathbf{H}}_t$
    \State $\mathbf{z}_t \leftarrow \mathrm{LayerNorm}(\mathbf{W}_{\mathrm{proj}} \tilde{\mathbf{H}}_t) \in \mathbb{R}^{D_{\mathrm{fuse}}}$
\EndFor
\Statex \textbf{Phase 3: temporal aggregation and scoring (RoTN)}
\State $\mathbf{Z} \leftarrow [\mathbf{z}_1, \dots, \mathbf{z}_T]^\top \in \mathbb{R}^{T \times D_{\mathrm{fuse}}}$
\State $\mathbf{Z} \leftarrow \mathbf{Z} + \mathbf{PE}_{1:T}$ \Comment{sinusoidal positional encoding}
\State $\mathbf{U} \leftarrow \mathcal{T}_\theta(\mathbf{Z})$ \Comment{causal mask online; KV cache reused per step}
\State $\mathbf{s} \leftarrow \mathrm{Classifier}(\mathbf{U}) \in \mathbb{R}^{T}$;\quad $\mathbf{a} \leftarrow \sigma(\mathbf{s}) \in [0,1]^{T}$
\Statex \textbf{Phase 4: GPU-accelerated temporal smoothing}
\State $\tilde{\mathbf{P}} \leftarrow \mathbf{a} \otimes \mathbf{1}_{24} \in \mathbb{R}^{T \times 24}$ \Comment{expand to frame resolution}
\State $\mathbf{P} \leftarrow \mathrm{Conv1D}\big(\mathrm{Pad}(\tilde{\mathbf{P}}),\, \mathcal{G}^{+}_\sigma\big)$ \Comment{causal half-kernel in online mode}
\State \Return $\mathbf{P}$
\end{algorithmic}
\end{algorithm}

\clearpage

\section{Limitations}
\label{app:limitations}

We state four limitations of the present study explicitly.

\paragraph{The dense comparison is a cost reference, not a controlled one.}
Table~\ref{tab:eff} places a dense 27B backbone in the same pipeline, where it runs
$2.24\times$ slower at $10.0\times$ the per-step FLOPs. Two things it does not settle.
Total capacity is 27B against 35B, so the pair is close to but not exactly matched. And a
dense backbone has no routing to read, so RoMF is inapplicable and its accuracy column
reflects a hidden-only readout; the accuracy gap therefore cannot be charged to sparsity
alone.

\paragraph{Backbone strength and routing modulation are not fully separated.}
The full-data baselines in Table~\ref{tab:main} use I3D, C3D or CLIP features, whereas RoMod reads a
frozen 35B-scale MoE. The ablations of Table~\ref{tab:ablation} and Appendix~\ref{app:fusion} isolate
the contribution of RoMF and RoTN \emph{given} this backbone, and Table~\ref{tab:eff} shows the effect
persists across three backbones, but a same-backbone baseline trained on $100\%$ of the weak labels
would be needed to attribute the margin over prior work between representation quality and routing
modulation.

\paragraph{The routing-only probe is already strong in-domain.}
Table~\ref{tab:routing-only} shows that a linear head on $\mathbf{R}_{\mathrm{clip}}$ alone reaches
$88.14\%$ AP on XD-Violence. The full framework adds $4.35$ AP on that benchmark. The larger part of
RoMod's value is therefore not raw in-domain accuracy but the constraint that keeps the semantic
pathway load-bearing, which is what shows up under domain shift (Appendix~\ref{app:fusion}) and across
layers (Section~\ref{sec:exp-layer}). We report the routing-only number rather than omitting it because
it bounds how much of the result a trivial readout already achieves.

\paragraph{Several diagnostics consume test-set labels.}
$J(l)$ (Appendix~\ref{app:fisher}), $S^{(l)}_c$ and $\Phi^{(l)}_c$ (Appendix~\ref{app:alignment}), and
the expert ranking $z^{(l)}_e$ used to choose the knockout set (Appendix~\ref{app:knockout}) are all
estimated with ground-truth temporal boundaries. They characterize the frozen backbone and are never
read by the deployed system, whose only layer-selection signal is $G(l)$ on $\mathcal{D}_{5\%}$. Read
as mechanistic evidence they are valid; read as a selection procedure they are not available at
deployment time, and Appendix~\ref{app:layer_adaptive} exists to show the deployable substitute
suffices.

\section{Extended Qualitative Visualizations}
\label{app:vis}

To examine the behaviour of RoMod under diverse surveillance conditions, this section provides extended
qualitative comparisons against the tuning-free hidden-state baseline across normal footage and three
real-world anomaly categories (riot, fighting, explosion).

\begin{figure}[p]
\centering
\includegraphics[width=\linewidth,height=0.44\textheight,keepaspectratio]{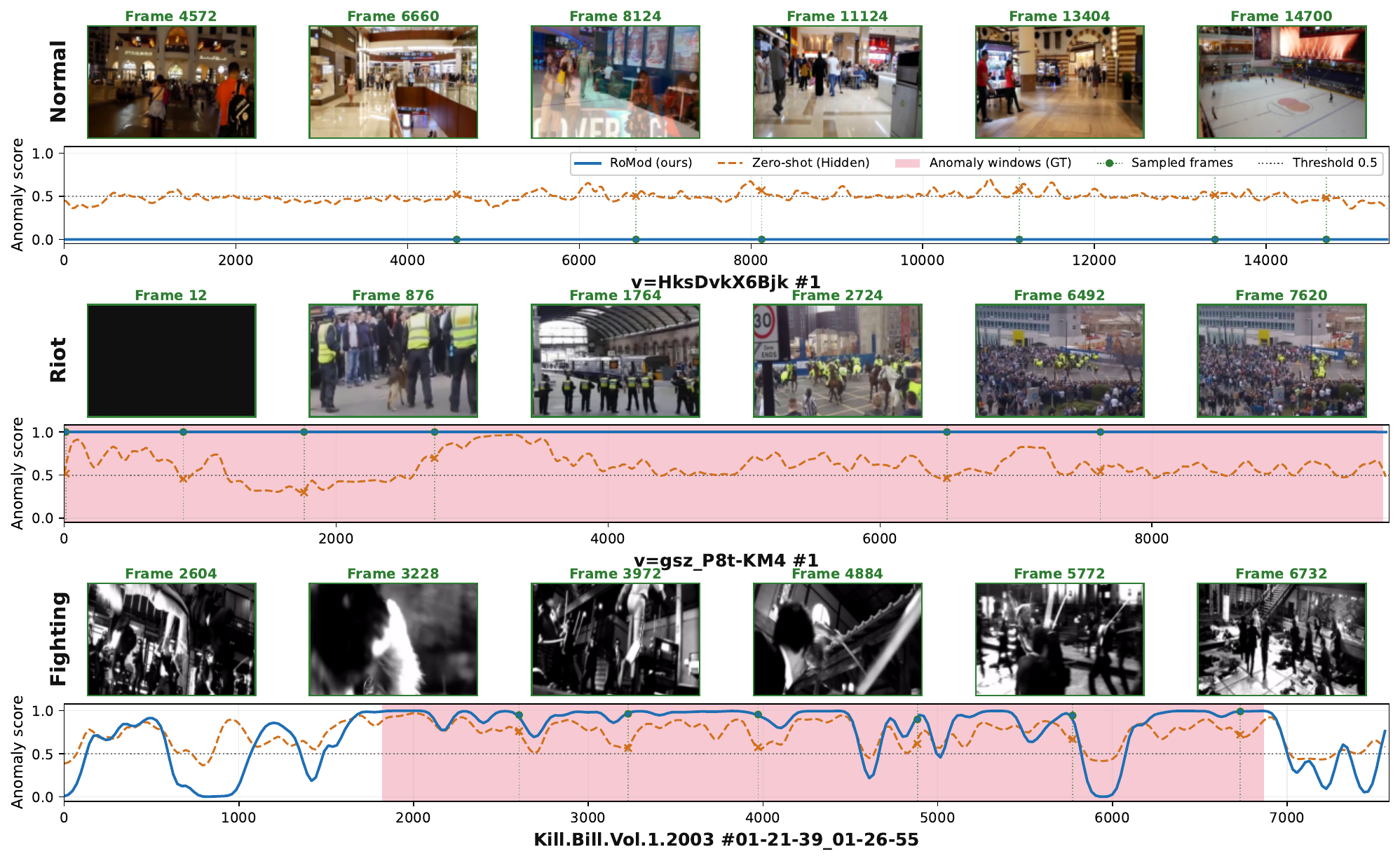}
\vfill
\includegraphics[width=\linewidth,height=0.44\textheight,keepaspectratio]{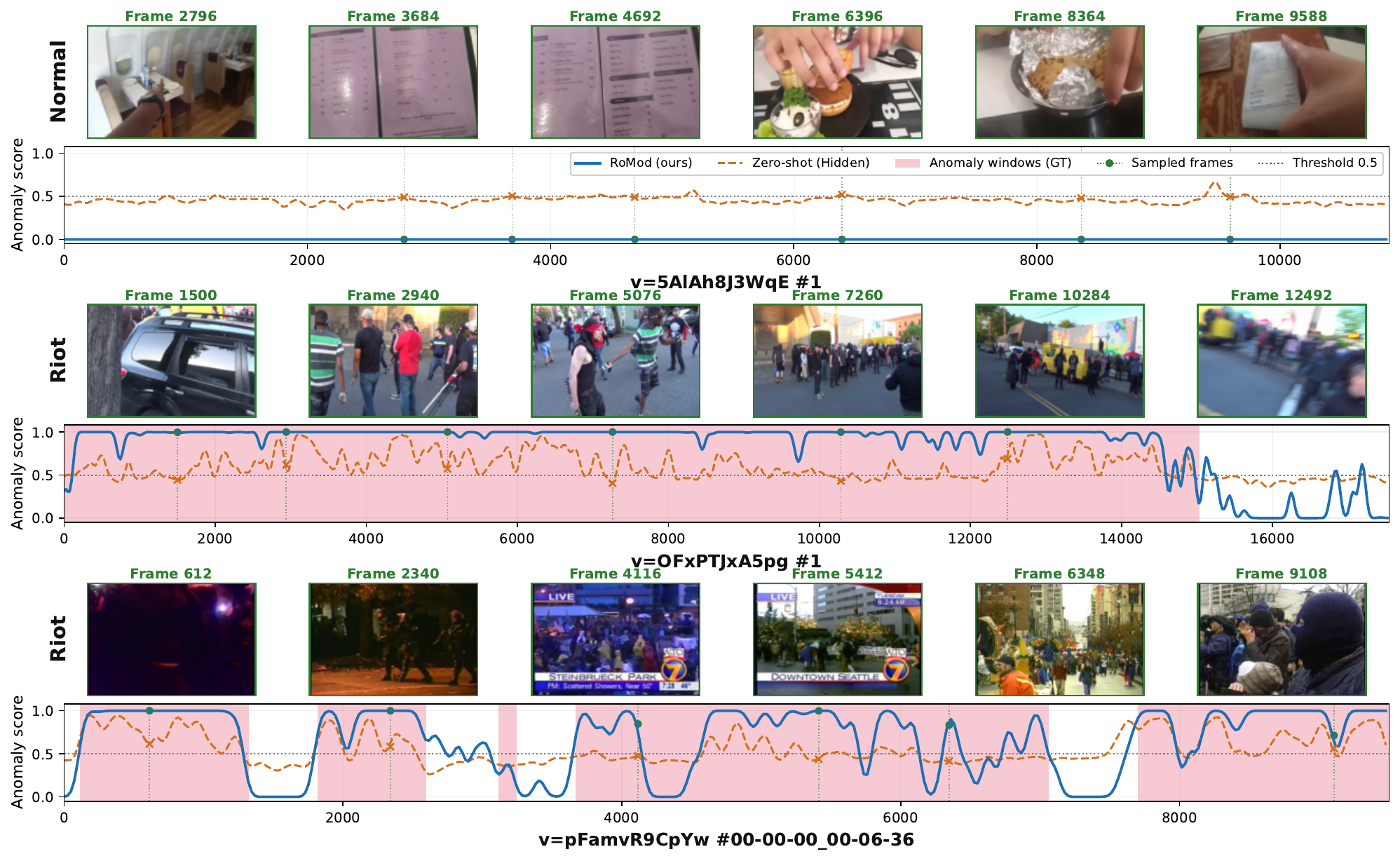}
\vspace{2pt}
\caption{\textbf{Extended qualitative anomaly localization curves on XD-Violence and UCF-Crime
(Part 1).}
\textbf{Top:} normal surveillance (\texttt{v=HksDvkX6Bjk}), long-range multi-segment riot
(\texttt{v=gsz\_P8t-KM4}), and abrupt motion transitions in fighting
(\texttt{Kill.Bill.Vol.1.2003}).
\textbf{Bottom:} robustness to illumination shifts in a normal video (\texttt{v=5AlAh8J3WqE}),
extended high-density crowd riot (\texttt{v=OFxPTJxA5pg}), and recurring civil unrest
(\texttt{v=pFamvR9CpYw}). Solid blue: RoMod; dashed orange: zero-shot hidden baseline; pink bands:
ground truth; dashed red: $\tau_{\mathrm{dec}} = 0.5$.}
\label{fig:vis_page1}
\end{figure}

\begin{figure}[p]
\centering
\includegraphics[width=\linewidth,height=0.44\textheight,keepaspectratio]{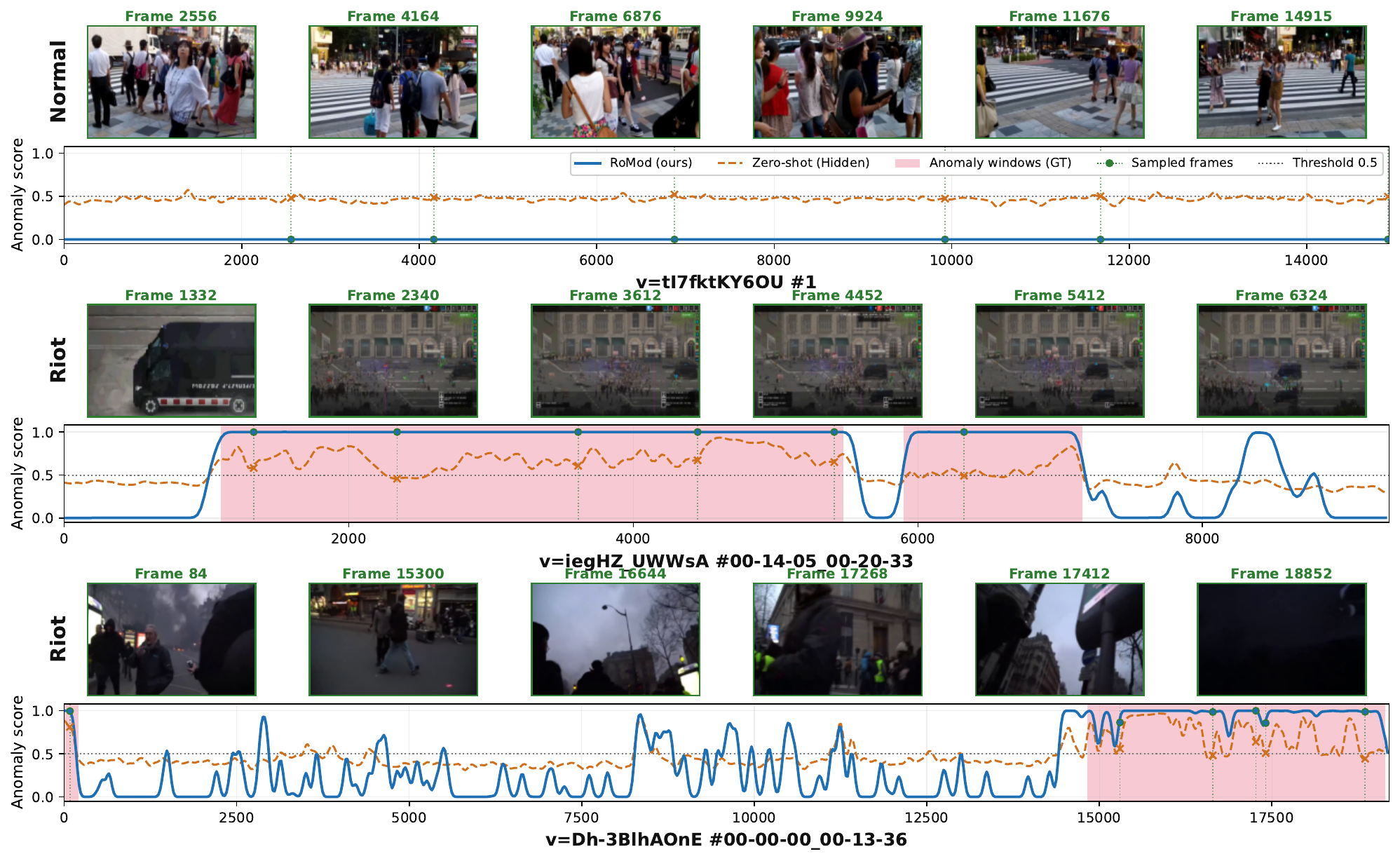}
\vfill
\includegraphics[width=\linewidth,height=0.44\textheight,keepaspectratio]{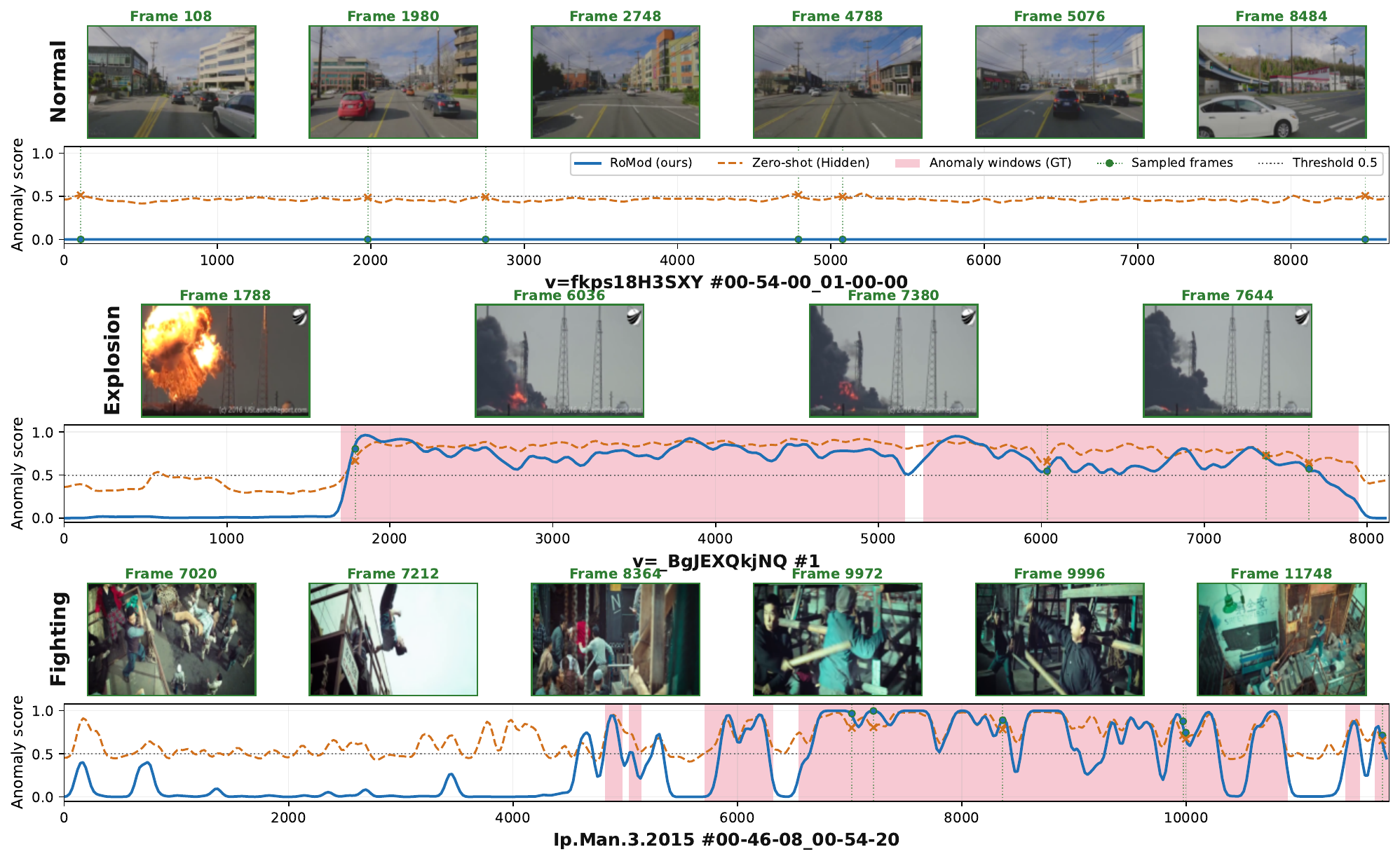}
\vspace{2pt}
\caption{\textbf{Extended qualitative anomaly localization curves on XD-Violence and UCF-Crime
(Part 2).}
\textbf{Top:} high-noise normal surveillance (\texttt{v=tI7fktKY6OU}), sustained multi-agent street
riot (\texttt{v=iegHZ\_UWWsA}), and ultra-long monitoring (\texttt{v=Dh-3BlhAOnE},
$> 18{,}000$ frames).
\textbf{Bottom:} ambient normal scene (\texttt{v=fkps18H3SXY}), transient vehicle explosion with rapid
smoke dissipation (\texttt{v=\_BgJEXQkjNQ}), and dense fighting interactions
(\texttt{Ip.Man.3.2015}).}
\label{fig:vis_page2}
\end{figure}

Figures~\ref{fig:vis_page1} and~\ref{fig:vis_page2} show continuous frame-level anomaly curves with
representative sampled keyframes for $12$ video sequences from XD-Violence and UCF-Crime. Ground-truth
anomalous segments are shaded pink, the decision threshold ($\tau_{\mathrm{dec}} = 0.5$) is the dashed
red line, RoMod is solid blue, and the zero-shot hidden probe is dashed orange.

\subsection{In-depth Qualitative Analysis}

Three behavioural regularities recur across the evaluated scenarios.

\paragraph{1. False-alarm suppression in complex normal footage.}
In purely normal sequences (\texttt{v=HksDvkX6Bjk}, \texttt{v=5AlAh8J3WqE}, \texttt{v=tI7fktKY6OU},
\texttt{v=fkps18H3SXY}) the zero-shot hidden baseline oscillates strongly. Camera panning, rapid
lighting shifts and dense non-anomalous pedestrian movement trigger spurious peaks that breach
$\tau_{\mathrm{dec}} = 0.5$ and frequently exceed $0.8$. RoMod holds the predicted score at or below
$0.1$ across all four videos. This is consistent with the account in Section~\ref{sec:method-romf}:
routing vectors in normal videos stay close to the global empirical baseline, so the multiplicative
residual $(\mathbf{1} + \boldsymbol{\gamma}_t)$ leaves background semantic channels unamplified.

\paragraph{2. Multi-interval boundaries in crowd riots.}
Long crowd-chaos sequences (\texttt{v=gsz\_P8t-KM4}, \texttt{v=OFxPTJxA5pg}, \texttt{v=Dh-3BlhAOnE})
are difficult because of the lulls between successive clashes. The uncalibrated hidden baseline decays
inside ground-truth anomalous intervals and recovers slowly during normal interludes, whereas RoMod
follows the multi-segment structure. Even in the ultra-long sequence exceeding $18{,}000$ frames
(\texttt{v=Dh-3BlhAOnE}) it tracks the entry and exit of each escalation and holds activations above
$0.9$ through sustained unrest.

\paragraph{3. Response to transient explosions and dynamic clashes.}
In a vehicle explosion (\texttt{v=\_BgJEXQkjNQ}) the anomaly occupies a few frames and is followed by
diffuse smoke and debris. The baseline lags, rising only after the visual explosion has peaked, while
RoMod steps up at the onset frame (frame $6036$) and saturates near $1.0$. In the high-velocity
confrontations of \texttt{Kill.Bill.Vol.1.2003} and \texttt{Ip.Man.3.2015} RoMod produces
near-rectangular contours matching the annotated fight windows, indicating that causal temporal
self-attention integrates history without smoothing away onset transients.

\clearpage

\end{document}